\documentclass[lettersize,journal,twoside]{IEEEtran}

\pdfoutput=1 
\usepackage{cite}
\usepackage{xcolor}
\usepackage{amsmath}
\usepackage{amssymb}
\usepackage{amsfonts}
\usepackage{graphicx}
\usepackage{algorithm}
\usepackage{algpseudocode}
\usepackage[hidelinks]{hyperref}
\usepackage{url}
\usepackage{dirtytalk}
\usepackage{enumitem}
\usepackage{tablefootnote}
\usepackage{threeparttable}

\usepackage{siunitx}
\usepackage{pgfplots}
\usepackage[font=footnotesize]{caption}
\usepackage{subcaption}
\pgfplotsset{compat = newest}

\begin{document}

\title{ClaPIM: Scalable Sequence CLAssification using Processing-In-Memory
%%% other suggestions
%ClassiPIM 
}

\author{\IEEEauthorblockN{Marcel Khalifa, \IEEEmembership{Student Member, IEEE},}
\and
\IEEEauthorblockN{Barak Hoffer,}
\and
\IEEEauthorblockN{Orian Leitersdorf, \IEEEmembership{Student Member, IEEE},}
\and
\IEEEauthorblockN{Robert Hanhan, \IEEEmembership{Student Member, IEEE},}
\and
\IEEEauthorblockN{Ben Perach, \IEEEmembership{Student Member, IEEE},}
\and
\IEEEauthorblockN{Leonid Yavits, \IEEEmembership{Member, IEEE},}
\and
\IEEEauthorblockN{and Shahar Kvatinsky, \IEEEmembership{Senior Member, IEEE}}

\thanks{This work was supported by the European Research Council through the European Union's Horizon 2020 Research and Innovation Program under Grant 757259. 
M. Khalifa, B. Hoffer, O. Leitersdorf, R. Hanhan, B. Perach, and S. Kvatinsky are with The Andrew and Erna Viterbi Faculty of Electrical \& Computer Engineering, Technion -- Israel Institute of Technology, Haifa 3200003, Israel (e-mail: \{marcelkh, barakhoffer, orianl, roberthanhan\}@campus.technion.ac.il; bperach@gmail.com; shahar@ee.technion.ac.il).

L. Yavits is with The Alexander Kofkin Faculty of Engineering, Bar-Ilan University, Ramat Gan 5290002, Israel (e-mail: leonid.yavits@biu.ac.il.). He is partially supported by European Union's Horizon Europe programme for research and innovation under grant agreement No. 101047160}
}

\maketitle

% ---- Abstract ---- %
\begin{abstract}

DNA sequence classification is a fundamental task in computational biology with vast implications for applications such as disease prevention and drug design.
%For example, the efficient and accurate identification of a pathogen may prevent the spread of disease and facilitate drug design.
Therefore, fast high-quality sequence classifiers are significantly important.
This paper introduces ClaPIM, a scalable DNA sequence classification architecture based on the emerging concept of hybrid in-crossbar and near-crossbar memristive processing-in-memory (PIM). We enable efficient and high-quality classification by uniting the filter and search stages within a single algorithm. Specifically, we propose a custom filtering technique that drastically narrows the search space and a search approach that facilitates approximate string matching through a distance function.
ClaPIM is the first PIM architecture for \emph{scalable} approximate string matching that benefits from the high density of memristive crossbar arrays and the massive computational parallelism of PIM. Compared with Kraken2, a state-of-the-art software classifier, ClaPIM provides significantly higher classification quality (up to $\boldsymbol{20\times}$ improvement in F1 score) and also demonstrates a $\boldsymbol{1.8\times}$ throughput improvement. Compared with EDAM, a recently-proposed SRAM-based accelerator that is restricted to small datasets, we observe both a $\boldsymbol{30.4\times}$ improvement in normalized throughput per area and a $\boldsymbol{7\%}$ increase in classification precision. 

\end{abstract}

% ---- Keywords ---- %

% ---- Introduction ---- %
\section{Introduction}
\label{sec:introduction}

\begin{figure}[t]
    \centering
    \includegraphics[width=\linewidth]{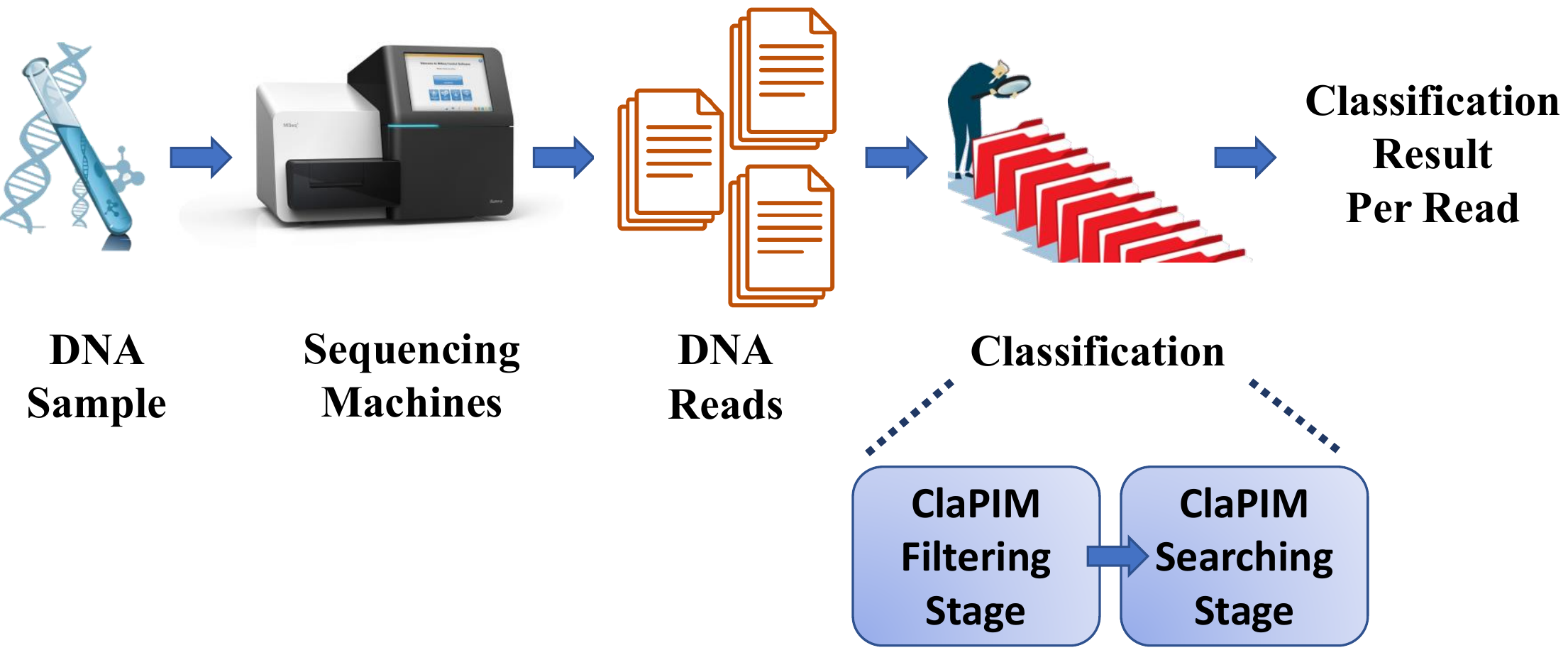}
    \caption{Overview of the overall taxonomic classification process, and the specific stages of the proposed ClaPIM classifier.}
    \label{fig:pipeline}
\end{figure}

Bioinformatics has significantly contributed to modern medical care through advances such as personalized medicine and accurate disease diagnostics. A fundamental task in bioinformatics is \emph{taxonomic DNA classification}: classifying genomes by species. This task is accomplished by first establishing a database of reference DNA sequences representing potential species and then sequencing a query DNA sample using \textit{sequencing machines} that produce numerous short \textit{reads} (sub-sequences of the sampled DNA). The goal of DNA classification is to determine the most likely species to which the sample belongs by comparing the reads to the database. As sequencing machines are prone to sequencing errors and given that mutations are common even between different samples of the same species, \textit{approximate sequence (string) matching} is advantageous when comparing reads to the reference DNA segments~\cite{AcceleratingGenomeAnalysis}. The classification pipeline is presented in Fig.~\ref{fig:pipeline}.

As opposed to the common case of exact string comparison, approximate string matching can tolerate edits up until a pre-defined \textit{edit distance} (number of substitutions, insertions, and deletions) between the two strings. Notice that approximate string matching also arises in other classification-based data-intensive applications such as data filtration for security monitoring, digital forensics, and data analytics~\cite{NIST}.

Unfortunately, traditional efficient solutions for string searching (e.g., suffix trees~\cite{suffix}) cannot support approximate matching. Other classification tools~\cite{MetaPhyler} implement approximate string matching by applying traditional optimal sequence alignment algorithms such as BLAST~\cite{BLAST}; these tools are slow since they are based on a dynamic-programming approach. Therefore, an efficient approximate heuristic-based approach known as $k$-mer matching ($k$-mer is a short DNA sub-sequence) is emerging in modern classifiers. The approach's idea is to extract $k$-mers from the DNA sequences and match the $k$-mers rather than the full sequences.
Classifiers include CLARK~\cite{CLARK}, LMAT~\cite{LMAT}, Kraken2~\cite{Kraken2}, KrakenOnMem~\cite{KrakenOnMem}, and Sieve~\cite{Sieve} are using this approach but the main drawback of these classifiers is that the $k$-mers are compared using exact matching, leading to high throughput at the cost of limited sensitivity.

Conversely, the recently introduced HD-CAM~\cite{HD-CAM} and EDAM~\cite{EDAM} accelerators proposes a heuristic that enables fast approximate comparisons of $k$-mers. The algorithm proposed by EDAM provides higher classification quality than HDCAM. Still, the main limitation of EDAM stems from the memory technology on which it is implemented: SRAM-based Content-Addressable Memory (SRAM-CAM). This memory technology suffers from poor density and performance scalability and thus cannot support large databases. Scalability is essential for sequence analysis infrastructures as databases on tens of billions of bases are common~\cite{Kraken}.

Since DNA classification is a data-intensive task that compares up to billions of sequences, processing-in-memory (PIM) platforms are highly suitable. Such platforms enable the data to be processed directly inside the memory without transferring it to a centralized processor. This enables PIM to overcome the \say{\textit{memory wall}}, a performance and energy bottleneck arising from data transfer between the processor and the memory. Memristive memories are an emerging form of PIM platforms as they inherently support data storage and computation within the same memory array~\cite{mMPU}.

This paper proposes ClaPIM, a novel taxonomic classification accelerator based on emerging PIM techniques. ClaPIM comprises \textit{filtering} and \textit{search} stages (Fig.~\ref{fig:pipeline}) that together enable efficient sequence classification. The search stage is inspired by an algorithm proposed in EDAM. Yet, the algorithm is modified to benefit from the massive density of memristive crossbar arrays (necessary for processing large databases). Furthermore, we propose an additional filtering step that drastically reduces the search space to increase energy efficiency and enable simultaneous in-memory comparisons for multiple DNA queries. Together, these solutions provide massive throughput on large datasets while retaining the superior quality provided by approximate $k$-mer matching.
Compared with Kraken2, a state-of-the-art \emph{exact-matching} software classifier, ClaPIM classification F1 score is $20\times$ greater while still providing a $1.8\times$ improvement in throughput.
Moreover, compared to EDAM, ClaPIM provides a $30.4\times$ improvement in normalized throughput per area (area efficiency) while also providing an improvement of up to $7\%$ in classification precision.

This paper makes the following contributions:
\begin{itemize}
    \item To the best of our knowledge, ClaPIM is the first scalable PIM-based accelerator for edit-distance-tolerant classification and is also the first taxonomic classification solution based on memristive memory.
    \item We propose a hybrid in/near-crossbar approach which exploits sensing circuitry towards efficient count and comparison operations.
    \item We propose a software filtering stage that improves both energy efficiency and classification precision on large datasets by $250\times$ and up to $7\%$, respectively.
    \item We propose a parallel query allocation scheme that increases throughput by dynamically activating different parts of the PIM architecture simultaneously.
\end{itemize}

% ---- Background ---- %
\section{Background}

\subsection{DNA Sequence Classification}
\label{sec:classification}

A DNA sequence is a string comprised of four basic molecules represented by the letters A, T, G, and C (known as bases).DNA sequence classification matches an unlabeled DNA sample to the closest sequence in an existing reference database. Different classification tasks, such as virus sub-typing or taxonomic classification, follow the same fundamental stages: (1) A metagenomic sample (mixture of DNA from multiple organisms and entities) is obtained, (2) the sample is inserted into a sequencer that outputs numerous small sequences of the DNA known as \textit{reads}, and (3) each read is compared against the existing database to associate it with known sequences. The sequencing errors carried by the reads, together with the mutations expected between samples of the same species, oblige classifiers to perform approximate string matching when evaluating the similarity between strings.

Classification quality is evaluated through a combination of the sensitivity and precision metrics, defined as follows:
\begin{equation}
    Sensitivity = \frac{TP}{TP+FN}, Precision=\frac{TP}{TP+FP},
\end{equation} 
where \textit{true positives (TP)} represents the number of correctly matched strings, \textit{false negatives (FN)} denotes the number of strings that should have matched but were not (potentially due to a sequencing error), and \textit{false positives (FP)} represents the number of strings that were falsely matched. Thus, 100\% sensitivity indicates that all true matches were identified, while 100\% precision indicates that all identified matches were correct. Since sensitivity trades off with precision and maximizing both metrics is desired, the F1 score is used to evaluate classification quality. This score is the harmonic mean of sensitivity and precision:

\begin{equation} 
F1=\frac{2 \times Sensitivity \times Precision}{Sensitivity+Precision}. 
\end{equation}

Traditional classifiers, e.g., BLAST-based models~\cite{MEGAN},~\cite{MetaPhyler}, apply sequence alignment algorithms to determine the similarity between strings. These alignment-based models are computationally expensive and time-consuming, and thus are limited when dealing with large-scale sequences. Alternatively, alignment-free tools have enabled high-throughput processing of sequencing data primarily due to their computational efficiency. The latter tools, such as Kraken2~\cite{Kraken2}, mainly exploit $k$-mer (read fragment of length $k$) exact matching heuristics; however, this results in reduced sensitivity in the presence of sequencing errors and mutations within $k$-mers.

HD-CAM~\cite{HD-CAM}, on the other hand, proposes tolerating Hamming distance to find the similarity regions. However, such technique mainly allows tolerating substitutions while providing little support to insertion and deletion tolerance.

EDAM~\cite{EDAM} resolves this shortcoming by tolerating edit rather than Hamming distance (i.e., supporting all types of edits). EDAM achieves edit distance tolerance by comparing each DNA basepair (a base) not only against the corresponding base in the query but also against its left and right neighbors. 
%This novel design allows for the application of a $k$-mer-based approximate matching algorithm.
Each $k$-mer from the database is stored in a CAM row. Each cell in the row stores one base. The query is fed to the CAM for comparison.
As the algorithm can tolerate edit distances (substitutions as well as insertions and deletions of bases), it provides a $30.9 \times$ improvement in sensitivity, and provides a $19.3 \times$ ($15.6 \times$) higher F1 score than Kraken2 (HD-CAM) for DNA sequences with high sequencing error rate.
However, the low scalability EDAM suffers from does not stem only from the low density of the memory technology. One of the apparent disadvantages of EDAM is that its cell is wire-bounded: every cell is physically wired to its right and left neighbors. This is a source of potential routing congestion and, therefore, more scalability limitations.

% EDAM~\cite{EDAM} proposed an edit-distance-tolerant approximate matching CAM design and a $k$-mer-based approximate matching algorithm.
% As the algorithm can tolerate edit distances (substitutions as well as insertions and deletions of bases), it provides a $30.9 \times$ improvement in sensitivity, and provides a $19.3 \times$ higher F1 score than Kraken2. 

\subsection{Classification Algorithm}
\label{sec:Classification Algorithm}

\begin{figure}
    \centering
    \includegraphics[width=\linewidth]{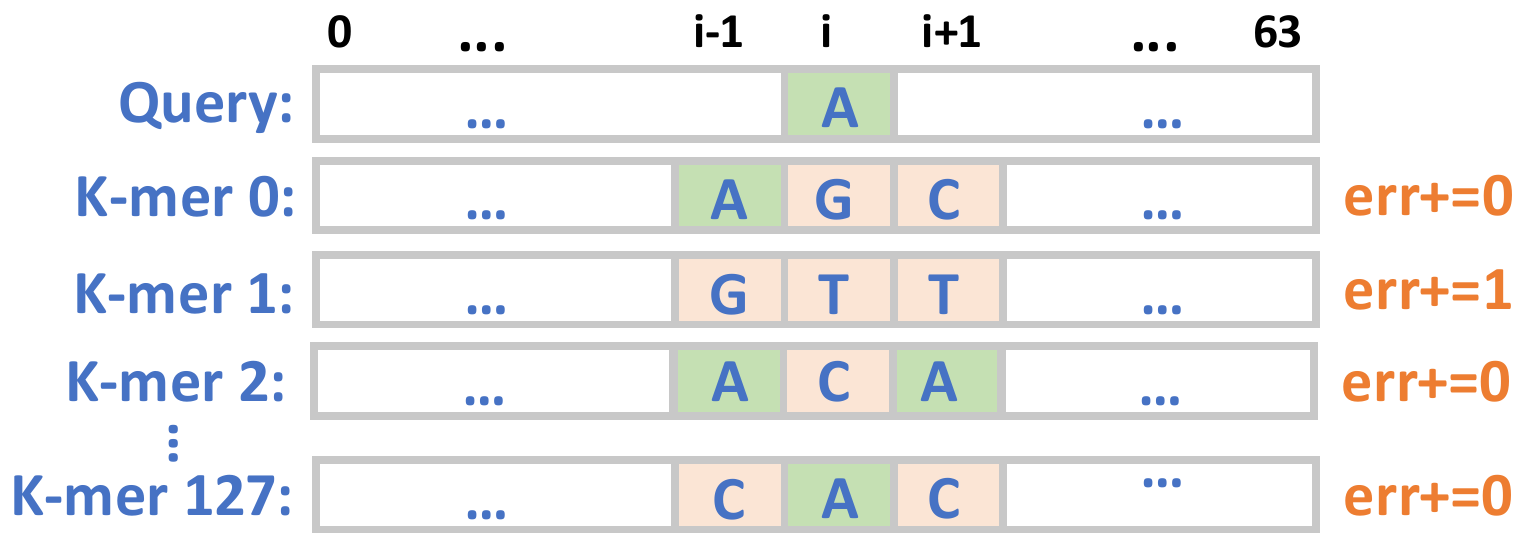}
    \caption{Comparing base $query[i]$ against co-located, left-neighbor and right-neighbor $k$-mer bases. If not matched, the error count is increased.}
    \label{fig:example}
\end{figure}

All $k$-mers (sub-sequences of length $k$) from different known genomes are pre-stored in a single database. ClaPIM assumes $k=64$; thus, for each DNA read from the sample, 64-mers are extracted and then queried against the database. The operating principle of EDAM's classification algorithm is based on the observation that an insertion or deletion shifts part of the data pattern to the right or left, respectively. Hence, by matching not only the co-located bases but also their left and right neighbors, it is possible to tolerate insertions and deletions, as illustrated in Fig.~\ref{fig:example}. If none of the three candidate data elements (co-located, left, and right neighbors) matches, a single element mismatch occurs, and one edit is counted. A query is considered a \textit{hit} against the $k$-mer only if the total number of edits does not exceed a predefined edit distance threshold. Lastly, the read is classified by selecting the genome with the maximum number of query hits for the read $k$-mers.

\subsection{Base-Count Filter}
\label{sec:Base-count Filter}

The base-count filter~\cite{filter} is a heuristic that aims to quickly estimate if the edit distance between two sequences exceeds $eth$, a predefined error threshold. FiltPIM~\cite{FiltPIM}, for example, uses the base-count filter as a filtering stage for the \textit{read alignment} task. The filter compares \textit{histograms} of two sequences $S_1$ and $S_2$. The number of occurrences of each base in $S_1$ is compared to the number of occurrences of the same base in $S_2$. For example, if $S_1$ has three more occurrences of \textit{A} than $S_2$, we infer that at least three edits exist. For each base $B$, we denote the number of its occurrences in the sequence $i$ as $B_i$.  Generally, two sequences are considered similar (the edit distance between them does not exceed $eth$) and will pass the filter only if:

 \begin{equation}
  |A_1 - A_2 | + |T_1 - T_2 | + |G_1 - G_2 | + |C_1 - C_2 | \leq 2eth.
 \end{equation}
 
\subsection{Processing within (Memristive) Memory}
\label{sec:PIM}
Data transfer between processing and memory units is among the leading factors limiting the performance, scalability, and energy efficiency of modern computing systems~\cite{wall}. PIM platforms may surmount this hurdle by uniting processing and memory units, especially for data-intensive applications. In this paper, we employ a memristive memory processing unit~\cite{mMPU} as the underlying PIM architecture, where memristor crossbar arrays both store the data and perform the computation. A memristor is an emerging nonvolatile memory technology that can store data in the form of resistance, logical `$0$' (`$1$') for high (low) resistance. This resistance is modified by applying a voltage across the memristor. ClaPIM utilizes two techniques for computation in/near the memory array: (1) digital stateful logic and (2) near-crossbar computing. While digital computing provides more accurate computing, near-crossbar computing is faster. We aim to benefit from both worlds.

\subsubsection{Stateful Logic}
\label{sec:MAGIC}

\begin{figure}
    \centering
    \includegraphics[width=\linewidth]{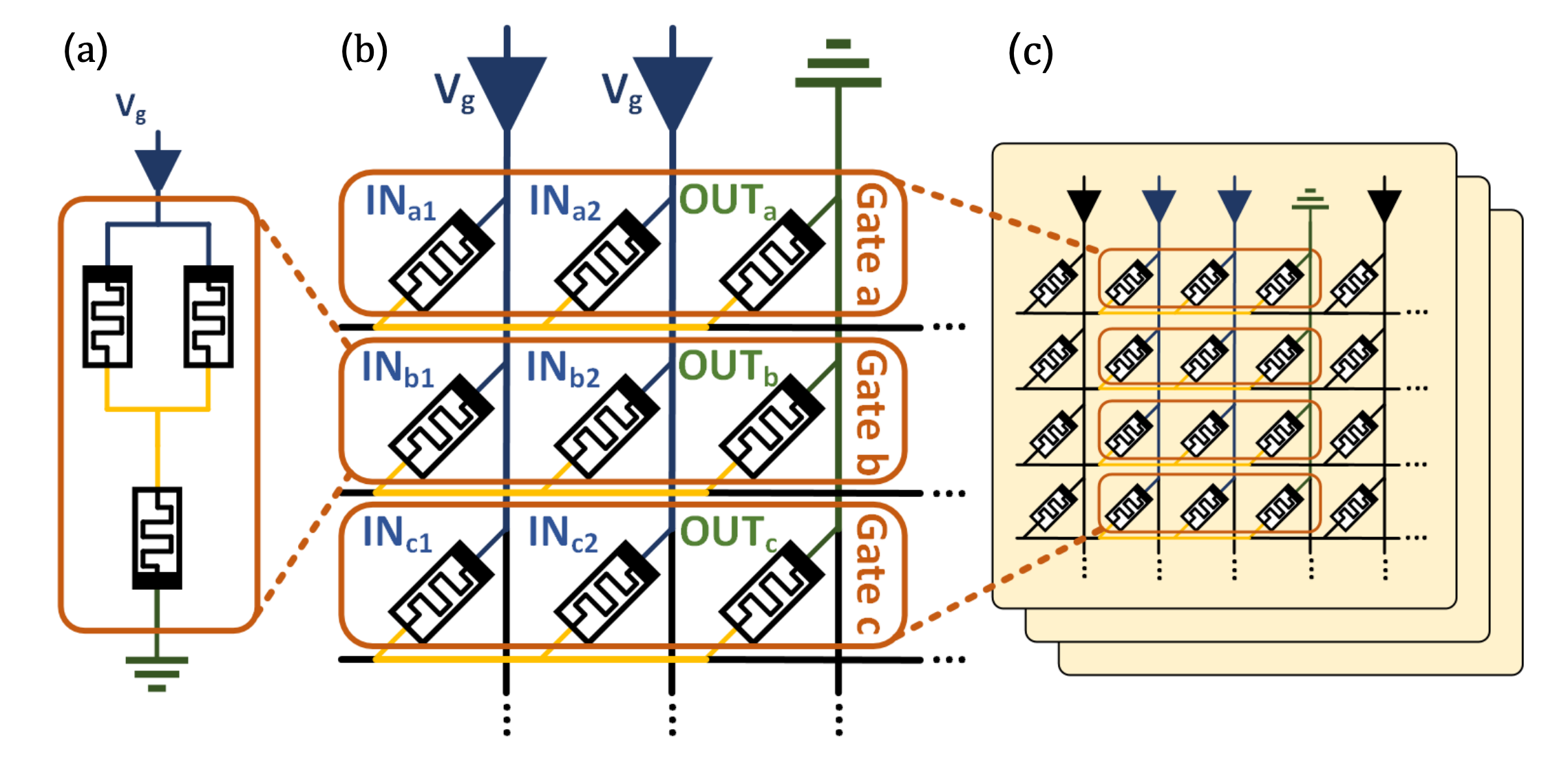}
    \caption{{(a) MAGIC NOR gate. (b) Parallel mapping of the MAGIC NOR gate to crossbar array rows, and (c) parallel computation across crossbars.}}
    \label{fig:XB}
    % \vspace{-15pt}
\end{figure}

An approach for memristor-based logic where the inputs and outputs of the gates are represented in the same form of data storage, e.g., resistance for memristors. A popular stateful-logic technique is Memristor Aided Logic (MAGIC)~\cite{MAGIC}, in which logic gates are performed on the resistances of the memristors directly inside the crossbar. The inputs of the MAGIC gate are the initial states (resistances) of the input memristors, and applying a fixed voltage $V_g$ to the memristor terminals results in the resistance of the output memristor after the computation storing the logical output. Fig.~\ref{fig:XB}(a) illustrates a MAGIC NOR gate as an example. The MAGIC NOR gate is crossbar compatible, as it can be performed using memristor cells within the same row of the crossbar by applying the voltages on crossbar bitlines. Moreover, MAGIC supports inherent parallelism as the same in-row gate can be performed in parallel across numerous rows (see Fig.~\ref{fig:XB}(b)) and across multiple crossbars (see Fig.~\ref{fig:XB}(c)). 

MAGIC NOR is executed in two steps (clock cycles): (1) initializing the output memristor to logical `$1$' (low resistance), (2) applying a voltage $V_g$ across the gate. Since NOR is functionally complete, all other functions can be performed using a sequence of MAGIC NORs. For example, in this work, we perform XOR using five MAGIC NOR cycles (not including initialization cycles, as they can be performed in parallel for all output memristors prior to the computation): 

\begin{equation}
    \begin{array}{l}
    a \oplus b = ((a' + b')' + (a + b)')'.
    \end{array}
\end{equation}

Other MAGIC (or different stateful logic) techniques can also be applied to implement XOR.
%in a different number of cycles but we focused on NOR-based MAGIC as it is a popular technique.}

\subsubsection{Near Crossbar Computing}
\label{sec:near crossbar}

\begin{figure}
    \centering
\begin{subfigure}{0.25\textwidth}
    \caption{}
    \includegraphics[width=\linewidth]{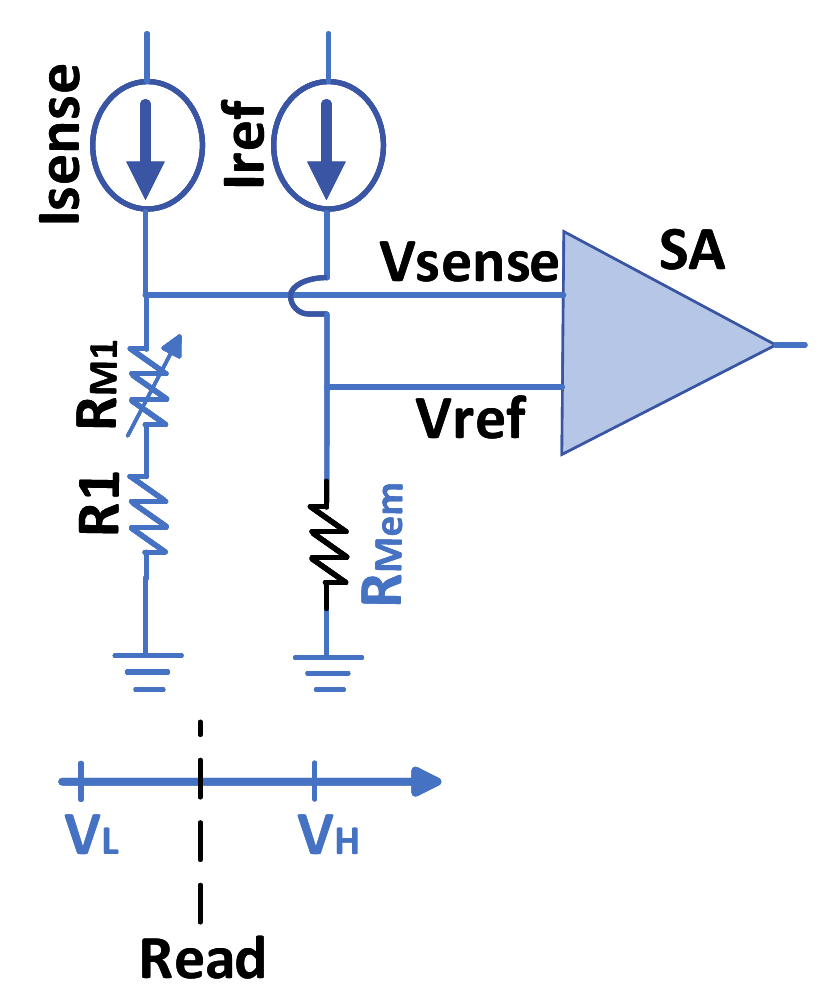}
    \label{fig:SA_read}
\end{subfigure}%
\begin{subfigure}{0.25\textwidth}
    \caption{}
    \includegraphics[width=\linewidth]{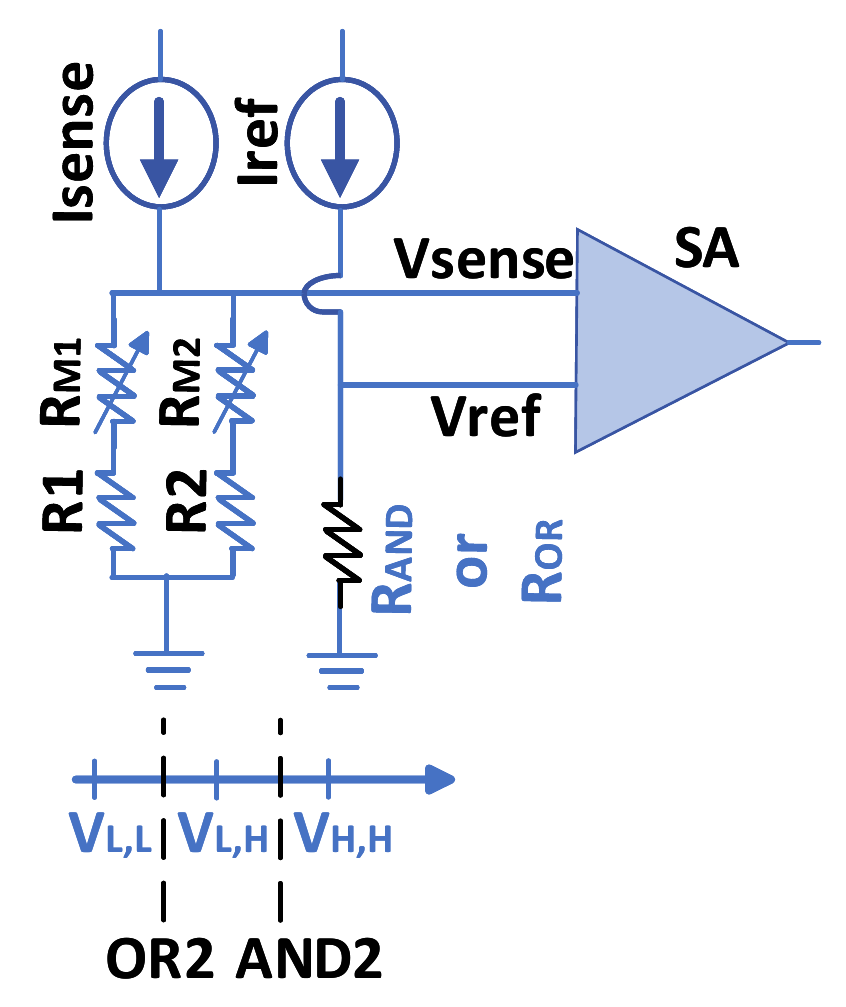}
    \label{fig:SA_op}
\end{subfigure}
    \caption{Voltage comparison between $V_{sense}$ and $V_{ref}$ for (a) memory read and (b) AND2/OR2 operation. }
    \label{fig:nearXB}
\end{figure}

% A computing model which requires data movement to the peripheral circuit (e.g., for state conversion) during the computation~\cite{PATMOS}. One technique to perform bulk bit-wise near-crossbar operation is to select the cell operands located in the same word-line. Then, comparing the equivalent resistance of the connected cells with a programmable reference by a sense amplifier (SA). This technique enables performing different basic operations like AND, OR and MAj as explained in AlignS~\cite{AlignS}.

In this approach, data is transferred to the peripheral circuitry (e.g., for state conversion) during the computation~\cite{PATMOS}. One such technique enables bulk bitwise near-crossbar logic by performing a read operation with two (or more) columns activated simultaneously. By activating multiple memristors per word-line in parallel, we are essentially sensing the equivalent parallel resistance of the selected memristors. Then, the sense amplifier's (SA) reference is chosen according to the equivalent resistances to perform different basic operations like AND, OR, and MAJ, as explained in AlignS~\cite{AlignS} and illustrated in Fig.\ref{fig:nearXB}.

\section{ClaPIM Architecture}
\label{sec:ClaPIM}

This section proposes the comparison operation performed in each memristive crossbar and then extends the proposed architecture to the filtering stage. Finally, the overall architecture and the data flow are presented. We use $k$-mer and 64-mer interchangeably as $k=64$.

\subsection{Querying within a Memristive Memory Array}
\label{sec:Within a Memristive Memory Array}

%%%%%%%%%%%%%%%%%%%%%%%%%%%%%%%%%%%%%%%%%%%%%%%%%%%%%%%%%%%%%%%%%%%%%%%%%%%%%%%%%%%%%%%%

\begin{figure}[t]
    \centering
    \includegraphics[width=\linewidth]{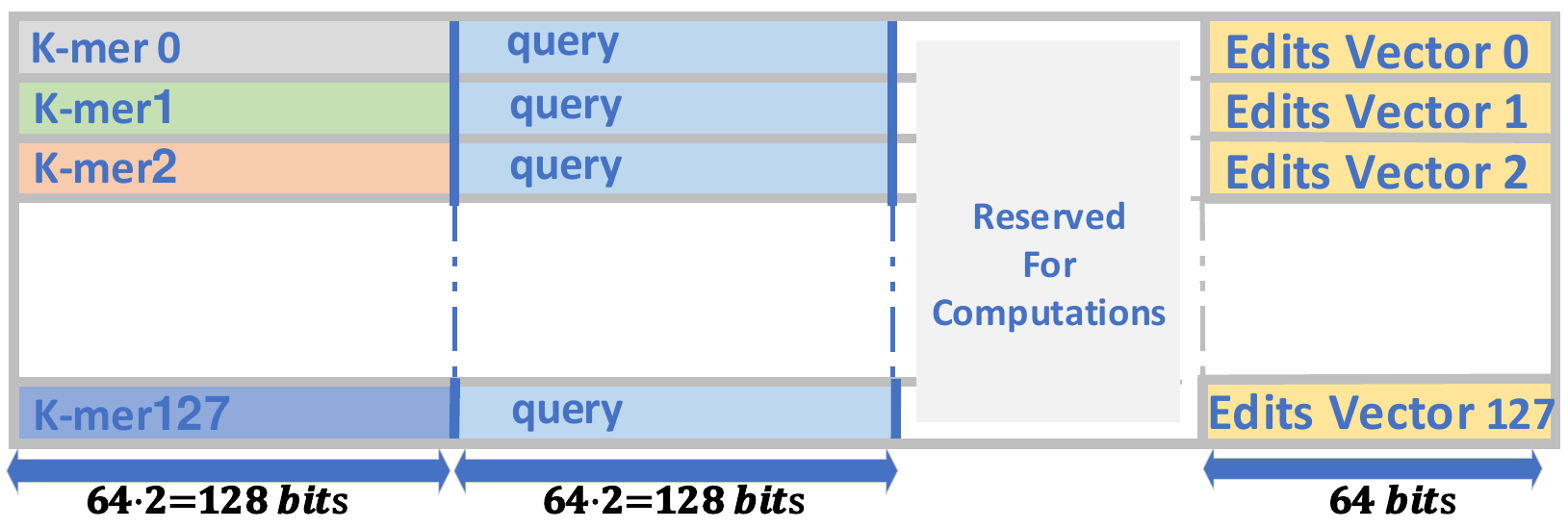}
    \caption{Mapping of data inside a 128$\times$512 memory crossbar array. The array is structured to support up to 64-mers (k-mers of length 64).}
    \label{fig:inMAT_FULL}
\end{figure}

%%%%%%%%%%%%%%%%%%%%%%%%%%%%%%%%%%%%%%%%%%%%%%%%%%%%%%%%%%%%%%%%%%%%%%%%%%%%%%%%%%%%%%%%

\begin{figure}
    \centering
\begin{subfigure}{0.5\textwidth}
    \caption{}
    \includegraphics[width=\linewidth,height=3cm]{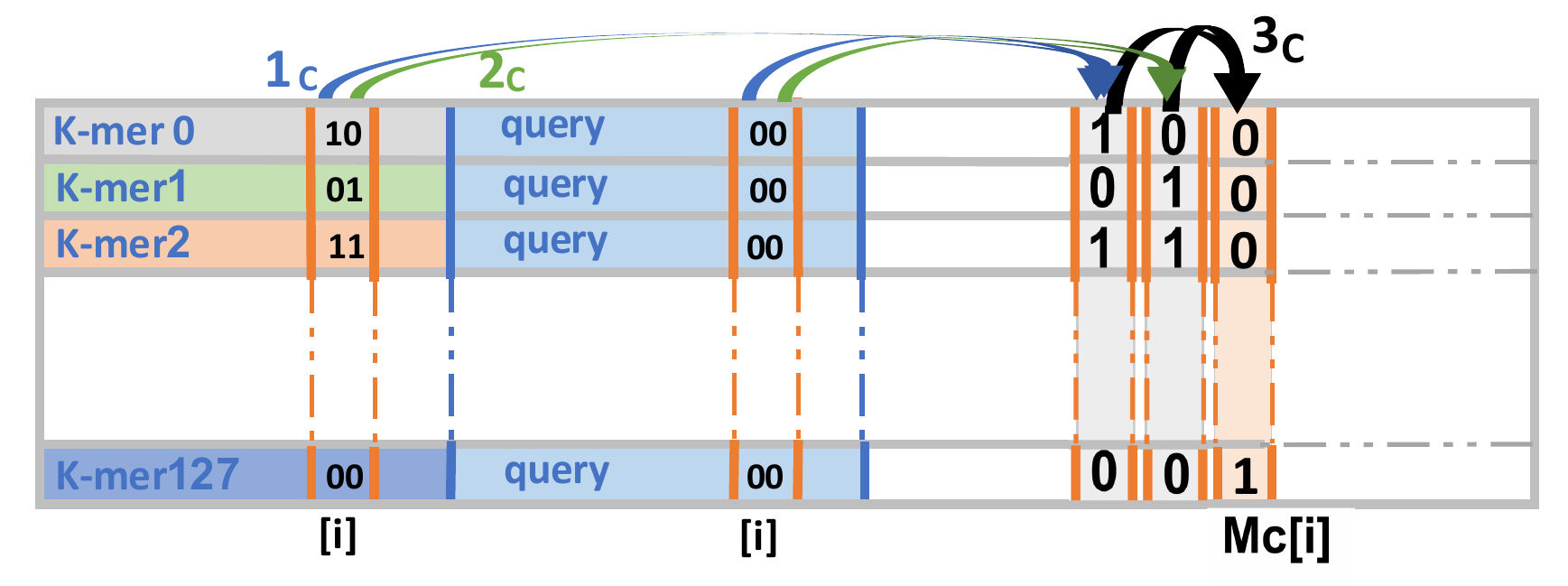}
    %\caption{In steps 1C and 2C, MAGIC XOR is performed for each bit to compare the query base against its co-located $k$-mer base. Then, in step 3C, MAGIC NOR is performed between the results of the two compares. Logical `$1$' in $M_C$ indicates the two bases matched.}
    \label{fig:inMAT_MC}
\end{subfigure}

\begin{subfigure}{0.5\textwidth}
    \caption{}
    \includegraphics[width=\linewidth,,height=3cm]{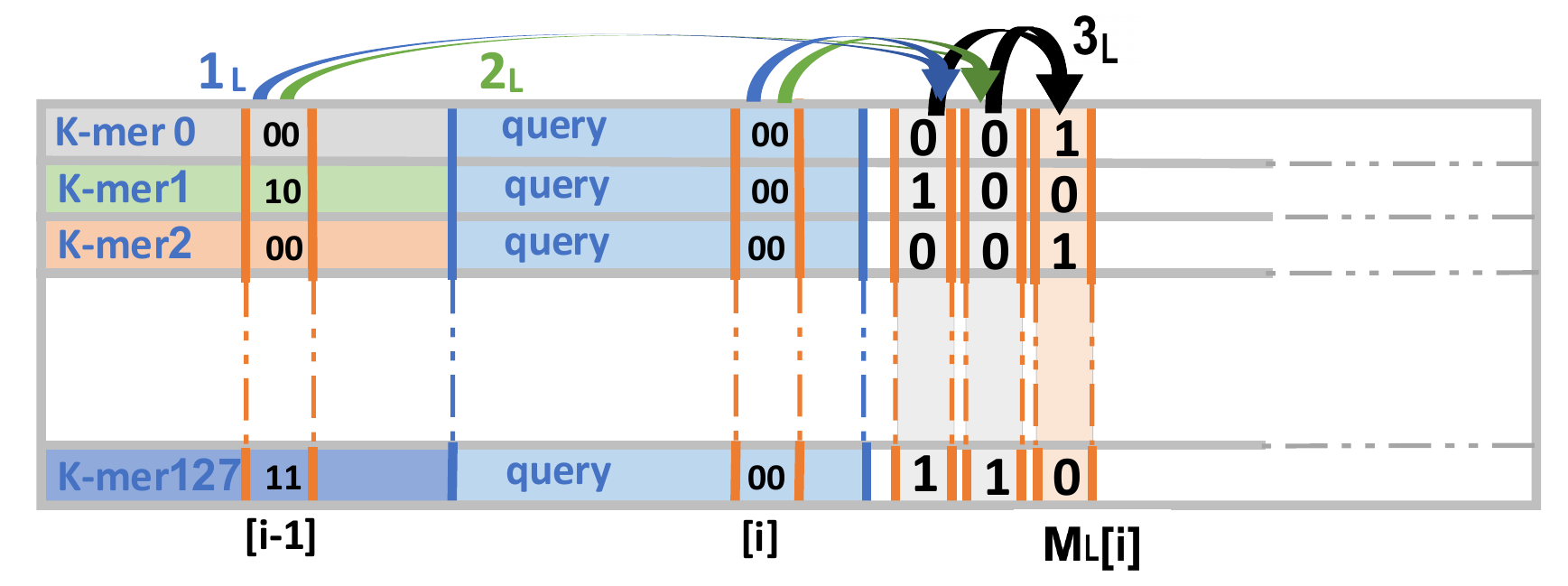}
    %\caption{In steps 1L and 2L, MAGIC XOR is performed for each bit to compare the query base against its left neighbor $k$-mer base. Then, in 3L, MAGIC NOR is performed between the results of the two compares. Logical `$1$' in $M_L$ indicates the two bases matched.}
    \label{fig:inMAT_ML}
\end{subfigure}

\begin{subfigure}{0.5\textwidth}
    \caption{}
    \includegraphics[width=\linewidth,,height=3cm]{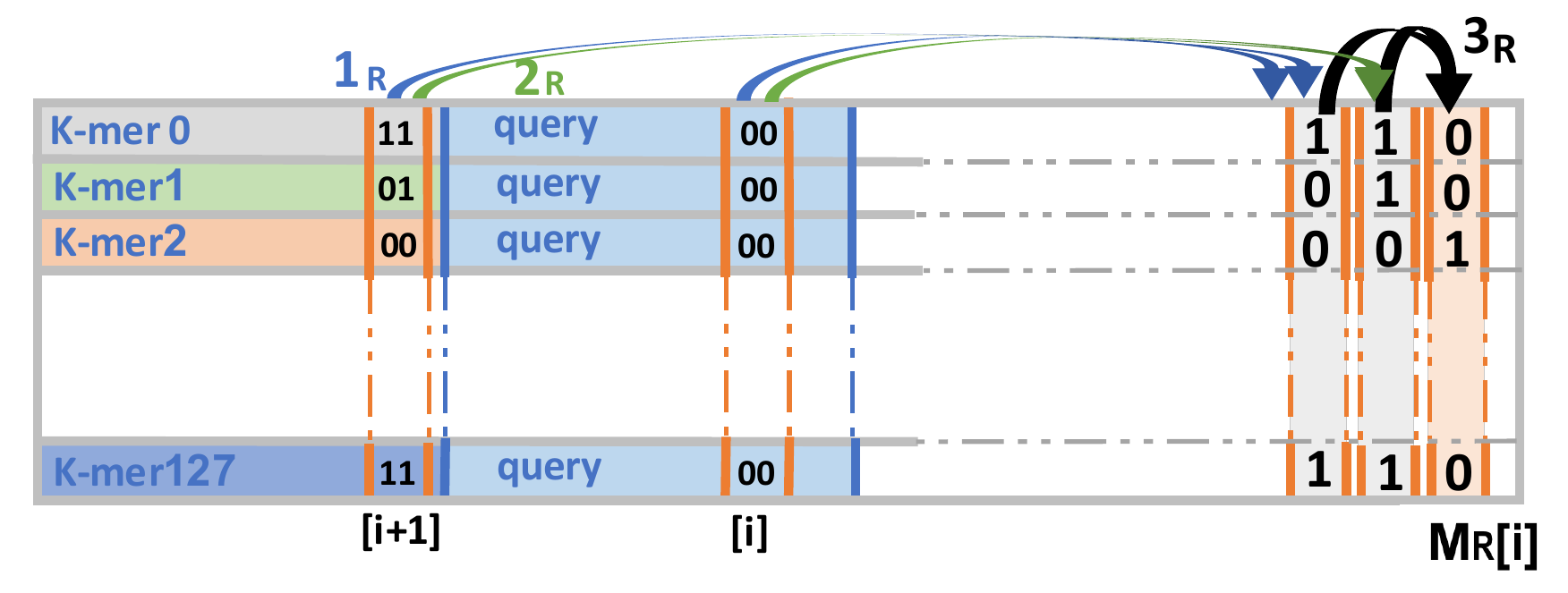}
    %\caption{In 1R and 2R, MAGIC XOR is performed for each bit, to compare the query base against its right neighbor $k$-mer base. Then, in 3C, MAGIC NOR is performed between the results of the two compares. Logical `$1$' in $M_R$ indicates the two bases matched.}
    \label{fig:inMAT_MR}
\end{subfigure}
    \caption{Comparing a query against the database within the memory crossbar. We denote $A=$ `00', $T=$ `01', $G=$ `10', and $C=$ `11'. The strings: \textit{Query}, \textit{$k$-mer} 0, \textit{$k$-mer} 1, \textit{$k$-mer} 2, ..., and \textit{$k$-mer} 127 are the same strings from Fig.~\ref{fig:example}. In steps 1\textit{X} and 2\textit{X}, XOR is performed for each bit, to compare the query base against its \textit{X} $k$-mer base. Then, in 3\textit{X}, MAGIC NOR is performed between the results of the comparison. Logical `$1$' in $M_X$ indicates the two bases matched, where \textit{X} is \textit{C} in (a) for the co-located base, \textit{L} in (b) for the left neighbor base and \textit{R} in (c) for the right neighbor base.}
    \label{fig:inMAT_M}
\end{figure}

%%%%%%%%%%%%%%%%%%%%%%%%%%%%%%%%%%%%%%%%%%%%%%%%%%%%%%%%%%%%%%%%%%%%%%%%%%%%%%%%%%%%%%%%

\begin{figure}[t]
    \centering
    \includegraphics[width=\linewidth]{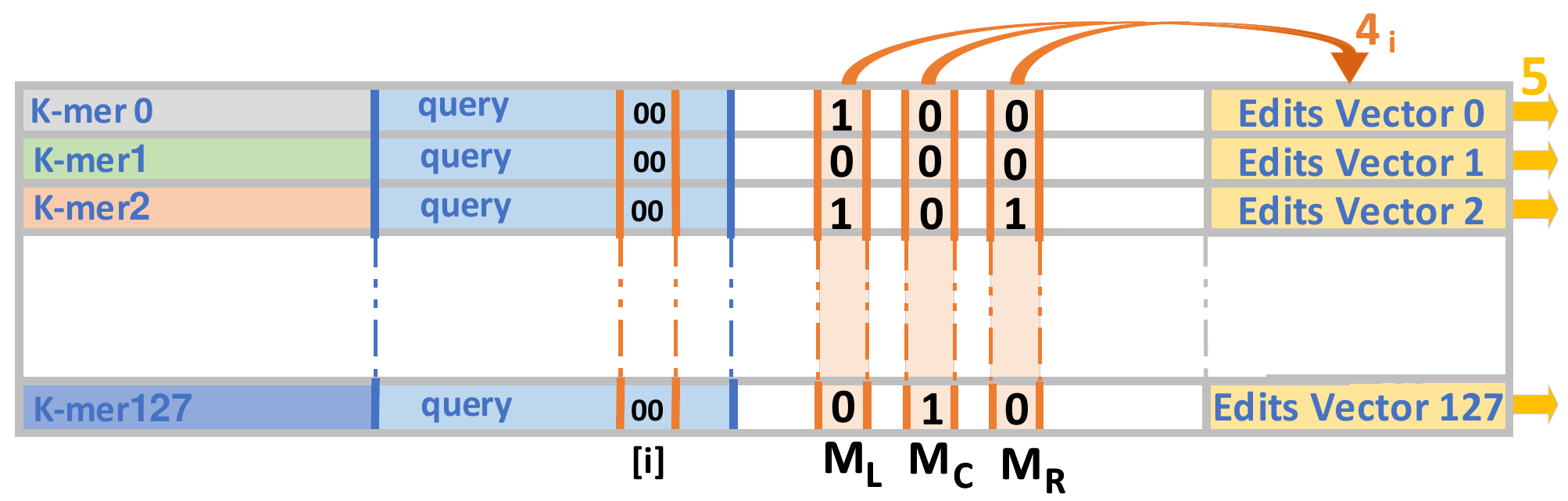}
    \caption{Step 4: As a three-input MAGIC NOR is performed, logical ‘1’ in bit $i$ in \emph{Edits Vector} indicates that the query base $i$ did not match any of the three bases to which it was compared (three horizontal zeros in MC[i], ML[i], and MR[i]); thus, it represents an edit.}
    \label{fig:inMAT_RES}
\end{figure}

The operating principle of ClaPIM's searching stage is adapted from EDAM's classification algorithm. We present the implementation of the search algorithm within a single memristive crossbar array. The array is structured to support up to 64--mers and queries; yet, the same principles, with the same data mapping technique, can be applied to larger crossbars to support higher values of $k$.

We consider 128$\times$512 arrays. 128 rows allow pre-storage of 128 64-mers from the database as each 64-mer is stored in a single row. To represent all four bases (A, T, G, C), two bits are required for each base. When a query arrives, it is written simultaneously to all rows. The data mapping inside the array is shown in Fig.~\ref{fig:inMAT_FULL}.

For each index $i$ in [0,63], we compare the two columns containing the query base $i$ to the two columns containing the corresponding $k$-mer base $i$. We use two MAGIC XOR gates to compare the two bits of the bases simultaneously in all rows (steps $1_C$ and $2_C$ in Fig.~\ref{fig:inMAT_MC}). We repeat this for the query base $i$ with respect to the left neighbor (right neighbor) $k$-mer base $i-1$ ($i+1$), through steps $1_L$ and $2_L$ ($1_R$ and $2_R$) in Fig.~\ref{fig:inMAT_ML} (Fig.~\ref{fig:inMAT_MR}).

A logical `$1$' in columns $M_C[i]$ ,$M_L[i]$, or $M_R[i]$ represents a match between the query base $i$ and its corresponding, left neighbor, or right neighbor base in the $k$-mer, respectively.
A match occurs only if both bits of the bases match, i.e., only if the result of both XORs is 0. Thus, to compute $M_C[i]$, a MAGIC NOR is executed (step $3C$); the same applies for $M_L[i]$ and $M_R[i]$.

\textit{Edits Vector} bits are calculated in steps $4_0,...,4_{63}$ serially, where in step $4_i$, a three-input MAGIC NOR gate is performed between the three columns ($M_C[i], M_L[i]$ and $M_R[i]$) of the query base $i$ (Fig.~\ref{fig:inMAT_RES}). A logical `$1$' bit in \textit{Edits Vector} indicates that this query base did not match any of the three bases to which it was compared (three horizontal zeros in $M_C, M_L$ and $M_R$); thus, it represents a single edit. To fit all these computations in a single row, we reuse intermediate cells; therefore, to reset memristor states, initialization cycles are utilized.

\begin{figure}
    \centering
    \includegraphics[width=\linewidth]{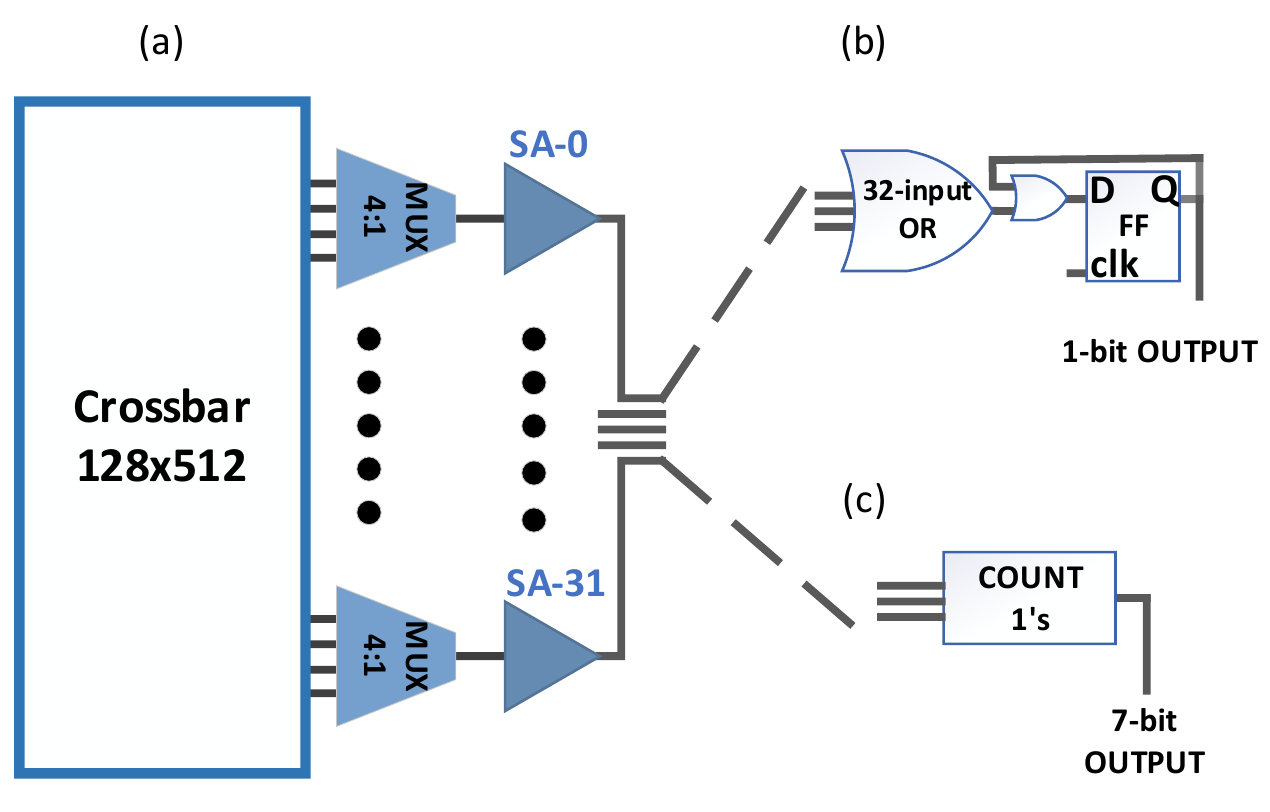}
    \caption{(a) The crossbar peripheries to perform the near-memory count and compare operations. The outputs of the \emph{Edits Vector} rows are connected to 32 SAs with the appropriate MUX circuits. The outputs of the SAs are either connected to (b) 32-input OR gate for detection tasks (since it does not matter how many matches are found as long as it is at least one), or to (c) a 1's counter for classification tasks (as the exact number of matches is required).}
    \label{fig:periphery3}
\end{figure}

\begin{figure}
    %\centering
    \includegraphics[width=\linewidth]{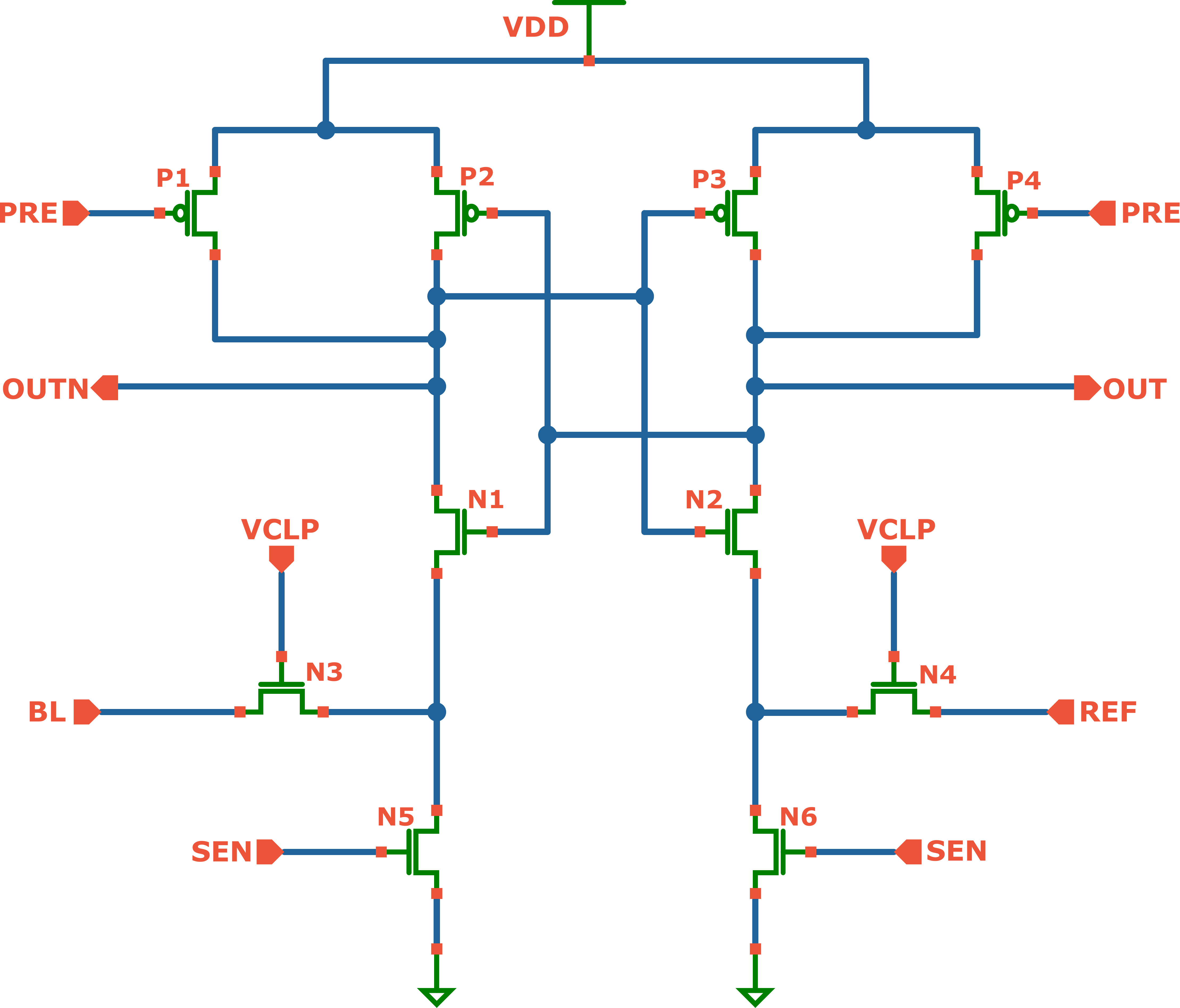}
    \caption{Schematic of the latched current sense amplifier (SA). OUT and OUTN are precharged to VDD using PMOS transistors (P1/P4), then the current of the selected BL and reference are compared by opening N3 and N4 with VCLP. The cross-coupled transistors (P2, N1/P3, N2) and pull-down NMOS transistors (N5/N6) push OUT and OUTN to VDD/GND and latch the result.}
    \label{fig:csa}
\end{figure}

Finally, in step $5$, the number of `$1$'s in each \textit{Edits Vector} is counted and compared to a certain threshold through a near-crossbar operation. We perform a single-cycle `read and count' operation by grounding all the bitlines and connecting a latched current-based sense amplifier (SA) to the wordline. Since all bitlines are activated simultaneously, we can measure the sum of their currents. The SA compares this sum to a reference current to determine if the number of bitlines storing logical `$1$' exceeds the threshold. The circuit design of the SA, shown in Fig.~\ref{fig:csa}, is based on~\cite{Lo2017}, with the reference circuit adjusted to support the required threshold. Hence, the required area for the SA is similar to that of a single-bit SA. To further reduce the area overhead, we use only 32 SAs for a 128$\times$512 crossbar, with the appropriate 4:1 multiplexing circuits, as shown in Fig.~\ref{fig:periphery3}a. To be able to process all rows, step $5$ is performed four times with 32 different rows selected each time.

For the general classification case, a single query can match different $k$-mers, in different crossbars belonging to other species. The species with the higher number of hits is selected to associate the query with a certain species. Therefore, we are interested in knowing if there is a hit or not and how many hits were found in each crossbar.  Therefore, the output of the SAs is fed into a \textit{counter} that produces and returns the total match count (Fig.~\ref{fig:periphery3}c). Suppose we are interested only in detection (detecting specific species in the sample) instead of classification. In that case, we only need to know if there is a hit in the reference database. Thus, the periphery can be reduced, with the \textit{counter} replaced by an OR gate and the crossbar returning only a 1-bit hit/miss (Fig.~\ref{fig:periphery3}b).
Finally, a simple network on chip in a tree-like topology is employed to gather the hit results from all the crossbars.

In summary, we propose parallel in-crossbar and near-crossbar operations that are efficiently used towards a query check against the $k$-mers stored inside the crossbar. 

\subsection{Filtering Stage}
\label{sec:Filter}

Naively performing the search for each query against all crossbars in massive datasets (up to tens of millions of crossbars) will lead to enormous energy consumption. Furthermore, this will shorten the lifetime of the limited write-endurance memory devices as they are frequently written.

Therefore, we introduce a filtering stage to produce a more efficient design. This stage is executed in the CPU before performing the search in the memristive memory. For each query, rather than comparing it against all crossbars, the query is compared only to crossbars containing $k$-mers with a \textit{potential hit} as defined by the filter. To find such $k$-mers, the \textit{histogram} of bases for each $k$-mer in the database is pre-computed offline, and all $k$-mers (of the same species) with the same histogram are stored together in sequential crossbars (if more than one crossbar is required). K-mers with different histograms reside in different crossbars. 
This manner of storing the k-mers hardly affects crossbar utilization. Applying this method when mapping the k-mers of chromosomes 1, 20, and 21, for example, resulted in crossbar utilization of $98.5\%$, $99\%$, and $98\%$, respectively.
Then, we store two numbers (start and finish indexes) for every potential $k$-mer histogram, specifying the range of its corresponding crossbars.

\begin{figure}
    \centering
    \includegraphics[width=\linewidth]{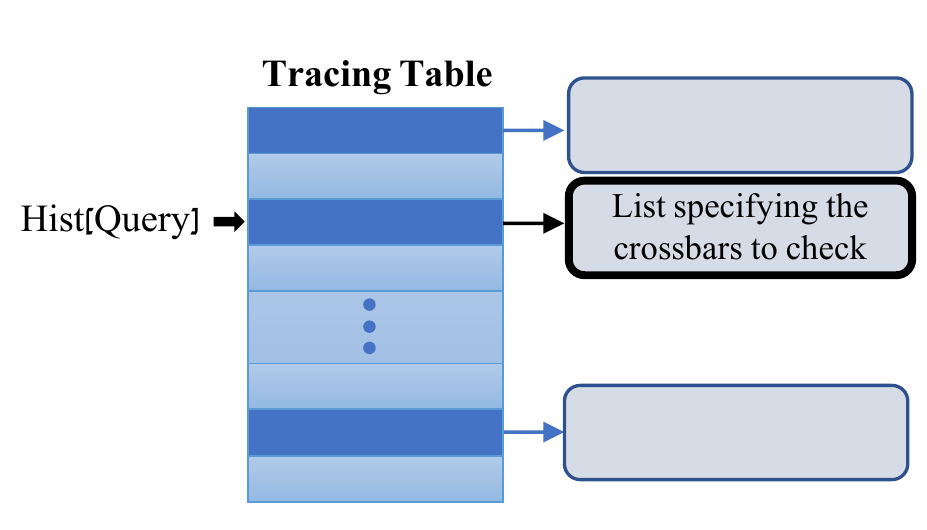}
    \caption{Extracting the ranges of the crossbars to be checked for the query. The dark blue rows in the table represent possible histograms for a query (there are a total of approximately 50K different possible histograms). The light blue rows are redundant rows representing histograms that will not appear (a histogram with zero $A$, $T$, $G$ and $C$ occurrences, for example). }
    \label{fig:tracing_table}
\end{figure}

During the operation, the histogram of each query is also computed. The query will only be compared to crossbars correlated with all its \textit{neighboring histograms}: histograms passing the base-count filter against our query histogram (see Section~\ref{sec:Base-count Filter}). To efficiently perform the \textit{tracing} of the crossbars with corresponding neighboring histograms, the processor maintains a \textit{tracing table} that contains, for each possible histogram, a pointer to a complete list of crossbar ranges belonging to the neighboring histograms, as illustrated in Fig.~\ref{fig:tracing_table}. 

There are approximately 50K different possible histograms for a string of length 64\footnote{Each histogram is represented by four integers whose sum is 64 (specifying the number of A, T, G, and C in the string). The number of possible histograms is equal to the combinatorial problem of distributing $64$ identical balls into four distinct boxes: $\binom{64+4-1}{4-1} $ $ \binom{67}{3}$ $=  47,905$.}.
The histogram representation is used as an index to access the tracing table.
The tracing table has $2^{18}$ rows as each histogram is represented by eighteen bits (six bits for \#A, \#T, and \#G while \#C can be calculated from these numbers). There are a total of $47,905$ different possible histograms and thus a total of $47,905$ lists.
By setting an \textit{eth} of, e.g., 4, each arbitrary histogram will have at most 309 neighboring histograms. Since three bytes will be enough to represent the largest crossbar index in the chip and two indexes are needed to represent each range Thus, the size of each list will be $309\cdot 2\cdot 3B = 1854B$. Thus, the table will use $2^{18} \cdot 4 bytes + 47,905 \cdot 1854 < 90MB$ of memory.

%By setting an \textit{eth} of, e.g., 4, each arbitrary histogram will have at most 309 neighboring histograms and, thus, the table will use less than 90MB of memory
%\footnote{The tracing table has $2^{18}$ rows as each histogram is represented by eighteen bits (six bits for \#A, \#T, and \#G while \#C can be calculated from these numbers). The histogram representation is used as an index to access the tracing table. There are a total of $47,905$ different possible histograms and thus a total of $47,905$ lists. The size of each list will be $309\cdot 2\cdot 3B = 1854B$ since three bytes will be enough to represent the largest crossbar index in the chip and two indexes are needed to represent each range. In total, $2^{18} \cdot 4 bytes + 47,905 \cdot 1854 < 90MB$.}.

The number of searches performed in each crossbar will be substantially reduced since the lookup in a crossbar is performed only if it has potential hits. This reduction will considerably increase the lifetime of the devices and reduce energy consumption as fewer devices are activated per query.

Additionally, the base-count filter boosts the precision of the classification algorithm since it can discard potential FP results, which the basic EDAM algorithm would have falsely matched. For example, let ``$CAC$" be the reference $k$-mer, ``$AAA$" be the query, and the edit distance threshold 1. The edit distance between the two strings is 2 (substitution-match-substitution); therefore, these two strings should generate a negative result (a mismatch). While EDAM would have considered it as a hit (thus generating an FP result), the base-count filter removes such $k$-mer from the list of potential hits, thereby improving the classification precision.

% \begin{figure*}
%     \centering
%     \includegraphics[width=\textwidth]{system1.pdf}
%     \caption{\textcolor{blue}{The mMPU chip architecture. The main modifications to conventional mMPU chip architecture are in the \textit{Read and Compute logic}  (dark blue). This block is presented and explained in Fig.~\ref{fig:periphery3}.}}
%     \label{fig:system}
% \end{figure*}

\begin{figure}
    \centering
    \includegraphics[width=\linewidth]{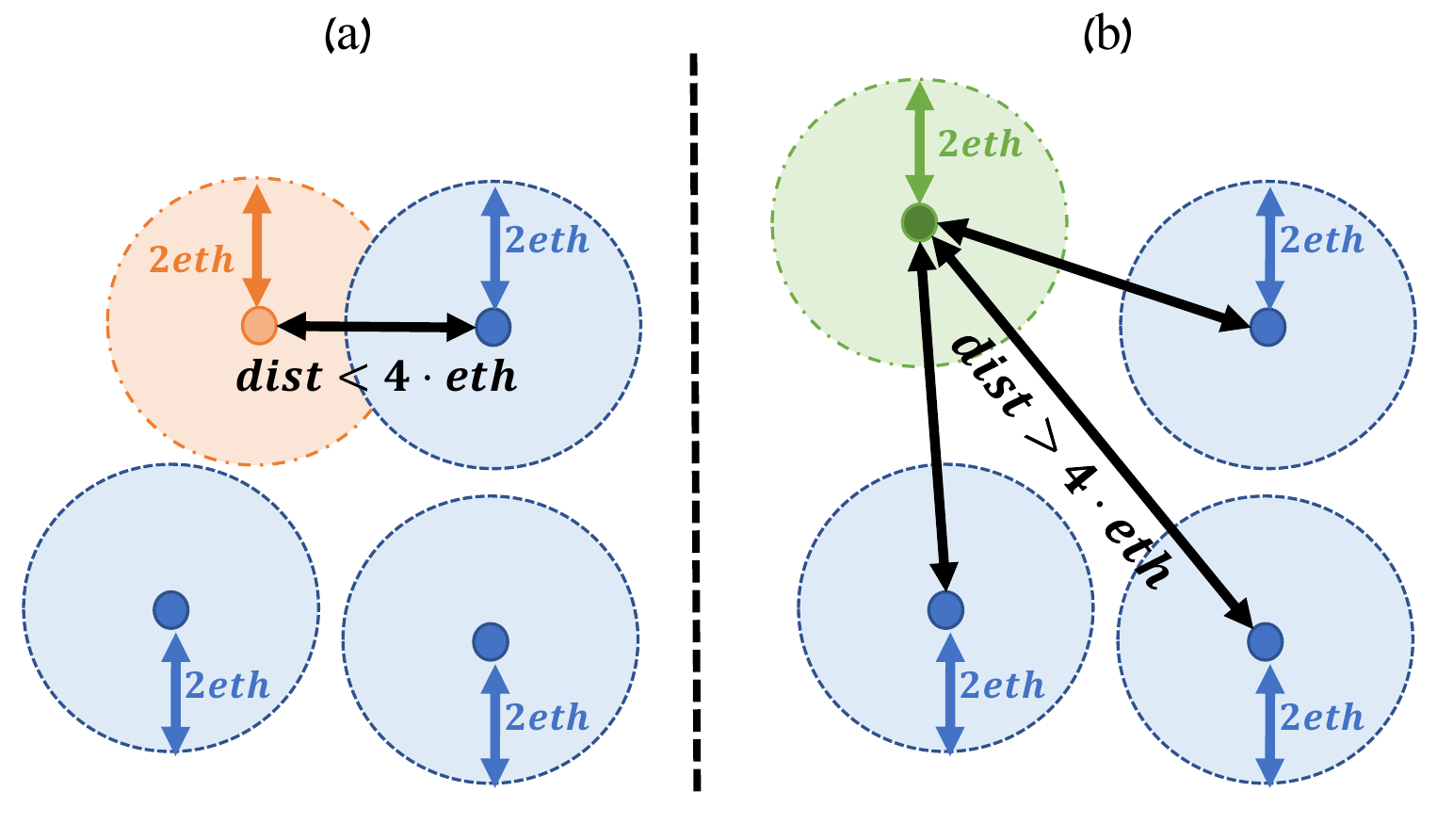}
    %\caption{Trying to add a new query to the batch. The center of a circle represents a histogram, while the actual circle contains all neighboring histograms of the same histogram. The centers of the blue circles represent histograms that are already in the batch. The orange center histogram in (a) is for a new query, that will not be added to the batch since the neighbors of its histogram overlap with other already selected histograms. The green center histogram in (b) is for a new query that will be added to the batch.}
    \caption{The filter aims to add a new query to the batch. First, the histogram of the new query is determined. The new query will be added only if the neighboring histograms of its histogram do not overlap with the neighboring histograms of already chosen queries in the batch. The center of the blue circles mark the histograms of queries in the batch, while the blue circle itself marks the neighboring histogram of such queries. In (a), the center of the orange circle is the histogram of a new query. This new query will not be added to the batch since its neighboring histograms overlap with already selected histograms. Conversely, in (b), the center of the green circle represents the histogram for a new query that will be batched with the others since there is no overlap between the neighboring histograms. }
    \label{fig:filter_batch}
    %\vspace{-15pt}
\end{figure}

Improvement in throughput can also be achieved by allowing different queries that access different crossbars to be checked simultaneously. To that end, we propose a \textit{batching} step on the CPU that dynamically allocates parallel queries. The algorithm proceeds as follows. We begin by adding the first query to the batch. Then, for each new query, if the query histogram is at least twice $ 2 \cdot eth $ apart from all histograms of the queries already in the batch, this new query will also be inserted. This condition guarantees that all of the chosen queries have non-overlapping neighboring histograms, meaning they access different crossbars and thus can be queried simultaneously. Fig.~\ref{fig:filter_batch} shows an example of adding two different queries with two different histograms to a batch. The center of each blue circle represents a histogram ($\#A, \#T, \#G, \#C$) of a query already in the batch while new queries are added. The orange center in (a) represents the histogram for a new query that will not be added to the batch since its neighboring histograms overlap with already selected histograms. On the other hand, the green center in (b) will be added to the batch as its neighboring histograms, which will be selected, do not overlap with the already selected histograms.

\subsection{ClaPIM: Putting it Altogether}
\label{sec:Full_Arch}

\begin{figure}
    \centering
    \includegraphics[width=\linewidth]{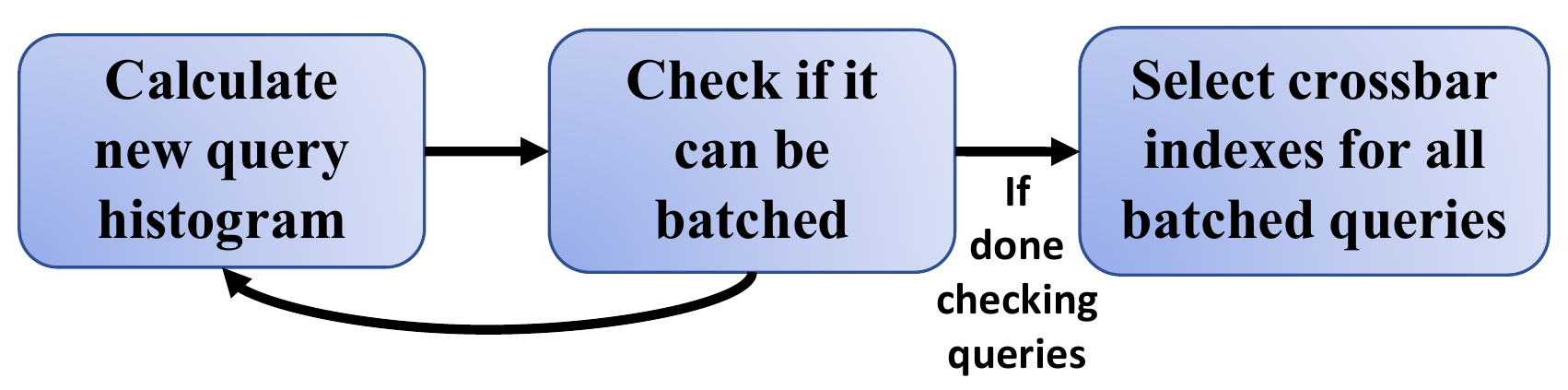}
    \caption{The overall flow of the filtering stage.}
    \label{fig:filtering}
\end{figure}
\begin{figure*}
    \centering
    \includegraphics[width=\linewidth]{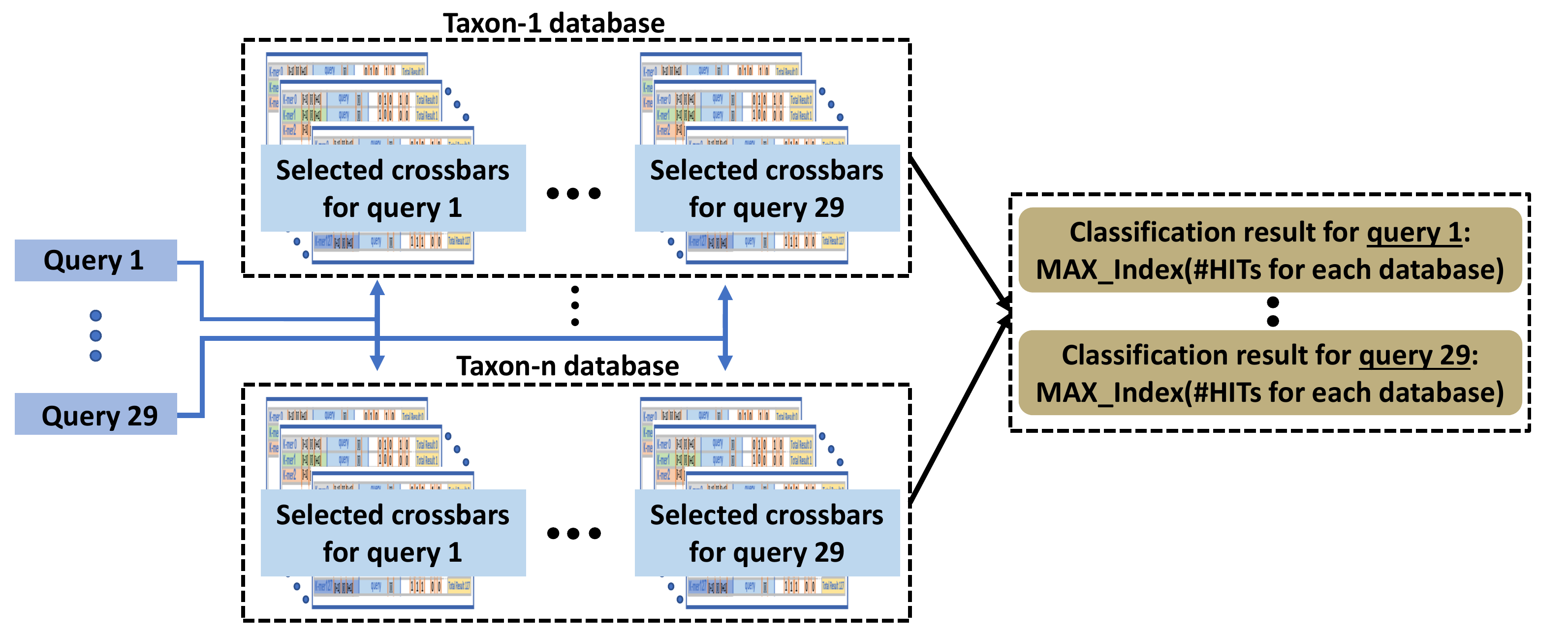}
    \caption{Overview of the PIM searching stage. Each query in the batch is searched for the selected crossbars of different taxons. Each query is associated with the database that provides the maximal number of hits for the query.}
    \label{fig:searching}
    %\vspace{-15pt}
\end{figure*}

Figs.~\ref{fig:filtering} and~\ref{fig:searching} present an overview of the ClaPIM filtering and searching stages, respectively. A memristive chip is assumed to be 8 GB in size (as suggested in~\cite{RACER}), containing 1M crossbars. Therefore, it can support a database of up to 128M 64-mers. For larger databases, several chips are used. Intel Optane, for example, uses DIMMs that embed up to 512GB~\cite{Optane}, thus enabling an 8G 64-mer database per DIMM.
% Fig.~\ref{fig:system} presents a high level overview for the mMPU chip architecture. Most of the peripheral circuits are adopted from prior work on mMPU (see~\cite{ref1}~\cite{ref2}). The primary modifications from the mMPU are the sense amplifier and near-memory logic as discussed in Section III-A.

The process starts with the \textit{batching} step as it simultaneously fills a list with different queries to be looked up in the chips. Then, for each query in the list, the range of crossbars to be compared against is extracted from the \textit{tracing table}, as explained in Section~\ref{sec:Filter}. These steps are performed in the CPU. The batching and tracing steps are combined in the filtering stage and pipelined with the PIM search stage. To ensure that the search operations completely mask the filtering time, we limit the number of queries that the batching step examines such that the run time should not exceed the run time of the searching stage.

The filtering time is independent of the number of chips used and will not change. Furthermore, the searching time will not change if more chips are used since the searching is done simultaneously in all chips. Thus, the limit for the number of queries checked in the batching size is independent of the number of chips used.

Finally, the memristive chips receive queries with different crossbar indices. The queries are serially written to their correlated crossbars, and only then the searching stage described in Section~\ref{sec:Within a Memristive Memory Array} starts in all crossbars for different queries simultaneously. For each query, the controller of the chip receives the number of hits against each reference organism (the sum of the number of hits received from all crossbars belonging to such an organism) and will classify the queries accordingly.  

\section{Evaluation}
\label{sec:Evaluation}

\begin{table}[t]
    \centering
    \caption{Memristor: Area, Power and Timing}
    \begin{tabular}{|c|c|c|}
        \hline
        \textbf{Attribute} & & \textbf{Source}\\
        \hline
        Cell Area & $9\cdot10^{-4} um^2$ & \cite{RACER} \\
        \hline
        MAGIC Cycle & $3ns$ & \cite{RACER} \\
        \hline
        Switching Energy & $6.4fJ$ & \cite{RACER} \\
        \hline
        SA Latency & $36ns$ & This work \\
        \hline
        SA Energy & $11.5pJ$ & This work \\
        \hline
        
    \end{tabular}
    \label{tab:Memristor}
    %\vspace{-12pt}
\end{table}

We now present the evaluation of ClaPIM. We begin with circuit evaluation to determine the latency and accuracy of the near-crossbar computing via the sense amplifiers (SAs). Then, the latency of the searching stage is determined to evaluate the maximally allowed latency for the filtering stage. The determination process is explained in the performance evaluation of the filtering stage. After that, the energy and lifetime improvements provided by the filtering stage are evaluated. Finally, the classification quality, throughput, and energy of ClaPIM are evaluated and presented.

\subsection{Circuit Evaluation}
\label{sec:circuit_eval}

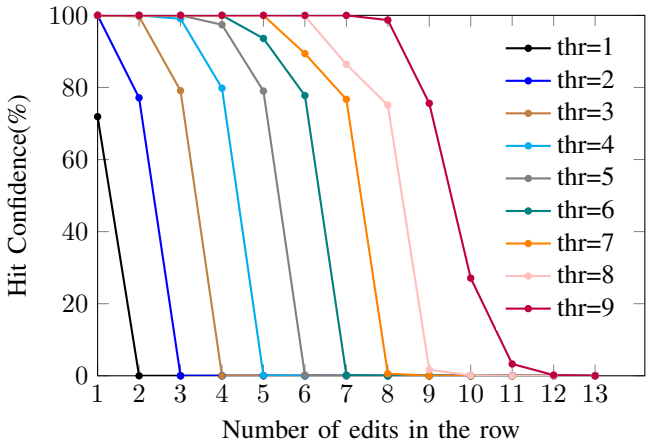
\begin{figure}[ht]
%\centering 
\begin{tikzpicture}
\begin{axis}[
    xlabel={Number of edits in the row},
    ylabel={Hit Confidence(\%)},
    xtick={1,2,3,4,5,6,7,8,9,10,11,12,13},
    ytick={0,20,40,60,80,100},
    ymin=0,
    ymax=100,
    xmin=1,
    legend style={draw=none},
    legend pos=north east,
    every axis plot/.append style={ultra thick},
    width=\columnwidth,
    height=0.72\columnwidth
    ]

\addplot[
    color=black,
    %smooth,
    thick,
    mark = *,
    mark size =1pt
    ]
    coordinates {(1,71.9)	(2,0)	(3,0)	(4,0)	(5,0)	(6,0)	(7,0)	(8,0)	(9,0)	(10,0)	(11,0)	(12,0) (13,0) 
};
\addplot[
    color=blue,
    %smooth,
    thick,
    mark = *,
    mark size =1pt
    ]
    coordinates {(1,100)	(2,77.13)	(3,0)	(4,0)	(5,0)	(6,0)	(7,0)	(8,0)	(9,0)	(10,0)	(11,0)	(12,0) (13,0) 
};
\addplot[
    color=brown,
    %smooth,
    thick,
    mark = *,
    mark size =1pt
    ]
    coordinates {(1,100)	(2,99.7)	(3,79.1)	(4,0)	(5,0)	(6,0)	(7,0)	(8,0)	(9,0)	(10,0)	(11,0)	(12,0) (13,0) 
};
\addplot[
    color=cyan,
    %smooth,
    thick,
    mark = *,
    mark size =1pt
    ]
    coordinates {(1,100)	(2,100)	(3,99.05)	(4,79.8)	(5,0.1)	(6,0)	(7,0)	(8,0)	(9,0)	(10,0)	(11,0)	(12,0) (13,0) 
};
\addplot[
    color=gray,
    %smooth,
    thick,
    mark = *,
    mark size =1pt
    ]
    coordinates {(1,100)	(2,100)	(3,100)	(4,97.4)	(5,79)	(6,0.2)	(7,0)	(8,0)	(9,0)	(10,0)	(11,0)	(12,0) (13,0) 
};
\addplot[
    color=teal,
    %smooth,
    thick,
    mark = *,
    mark size =1pt
    ]
    coordinates {(1,100)	(2,100)	(3,100)	(4,100)	(5,93.6)	(6,77.76)	(7,0.2)	(8,0)	(9,0)	(10,0)	(11,0)	(12,0) (13,0) 
};
\addplot[
    color=orange,
    %smooth,
    thick,
    mark = *,
    mark size =1pt
    ]
    coordinates {(1,100)	(2,100)	(3,100)	(4,100)	(5,100)	(6,89.4)	(7,76.7)	(8,0.6)	(9,0)	(10,0)	(11,0)	(12,0) (13,0)
};
\addplot[
    color=pink,
    %smooth,
    thick,
    mark = *,
    mark size =1pt
    ]
    coordinates {(1,100)	(2,100)	(3,100)	(4,100)	(5,100)	(6,99.8)	(7,86.4)	(8,75.1)	(9,1.7)	(10,0.1)	(11,0)	(12,0) (13,0) 
};
\addplot[
    color=purple,
    %smooth,
    thick,
    mark = *,
    mark size =1pt
    ]
    coordinates {(1,100)	(2,100)	(3,100)	(4,100)	(5,100)	(6,100)	(7,100)	(8,98.7)   (9,75.6)	(10,27.1)	(11,3.28)	(12,0.2) (13,0) 
};

\legend{thr=1,thr=2,thr=3,thr=4,thr=5,thr=6,thr=7,thr=8,thr=9}
\end{axis}
\end{tikzpicture}
\caption{Monte Carlo experiments for the memristive model variations. ``thr": edit distance threshold allowed to consider the row as a hit.}
\label{fig:MC}
%\vspace{-12pt}
\end{figure}

Table~\ref{tab:Memristor} lists the area, switching latency, and energy for the SA and the ReRAM memristive devices used in the design. The peripheral circuits were designed and evaluated with Cadence Virtuoso using the iHP SG13S process with MEMRES PDK, a fabrication-ready CMOS-ReRAM integrated process~\cite{IHPPDK}. To evaluate the correctness of the near-crossbar count and compare (performed using the SA), extensive Monte Carlo simulations, 1000 iterations, were employed. The evaluation considered the presence of device mismatches and process variation for both the memristors and transistors. Fig.~\ref{fig:MC} shows the hit confidence of the SA. For example, if the edit distance threshold (\textit{thr}) is four, then all rows containing up to four logical `$1$'s should be considered a hit. The graph shows that if the row contains three (five) logical `$1$'s, then there is a $100\%$ ($0\%$) chance of considering the row as a hit. Nevertheless, for a row containing exactly four logical `$1$'s, the SA will consider it a hit $79.8\%$ of the times. As an ideal SA would have provided a step function for each \textit{thr}, we conclude that our SA performs adequately.

%%%%%%%%%%%%%%%%%%%%%%%%%%%%%%%%%%%%%%%%%%%%%%%%%%%%%%%%%%%%%%%%%%%%%%%%%%%%%% subfigure 7 

%%%%%%%%%%%%%%%%%%%%%%%%%%%%%%%%%%%%%%%%%%%%%%%%%%%%%%%%%%%%%%%%%%%%%%%%%%%%%%%%%%%%%%%%%%%%
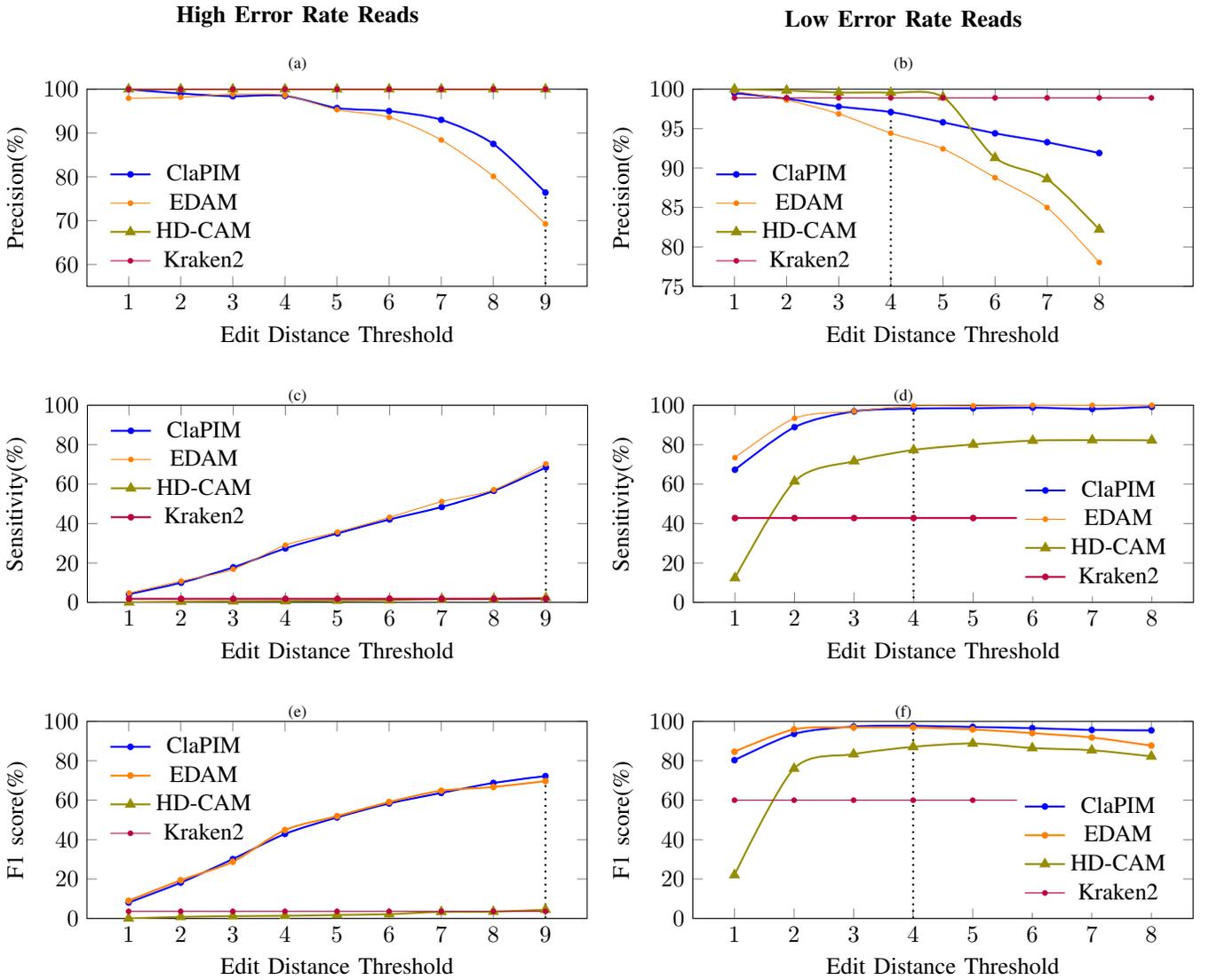
\begin{figure*}[!t]
    \centering
%%%%%%%%%%%%%%%%%%%%%%%%%%%%%%%%%%%%%%%%%%%%%%%%%%%%%%%%%%%%%%%%%%%%%%%%%%%%%% subfigure 1 
\begin{subfigure}{0.5\textwidth}
\centering
\textbf{High Error Rate Reads}\par\medskip
\caption{}%EDAM vs ClaPIM precision for high error rate reads.
\begin{tikzpicture}
\begin{axis}[
    xlabel={Edit Distance Threshold},
    ylabel={Precision(\%)},
    xtick={1,2,3,4,5,6,7,8,9},
    ytick={60,70,80,90,100},
    ymin=55,
    ymax=100,
    legend style={draw=none,fill=none},
    legend pos=south west,
    every axis plot/.append style={ultra thick},
    width=\columnwidth,  
    height=0.5\columnwidth,
]
\addplot[
    color=blue,
    smooth,
    thick,
    mark = *,
    mark size =1pt
    ]
    coordinates {(1,100)(2,99)(3,98.4)(4,98.5)(5,95.7)(6,95)(7,93)(8,87.5)(9, 76.4)};
\addplot[
    color=orange,
    smooth,
    thin,
    mark = *,
    mark size =1pt
    ]
    coordinates {(1,97.94)(2,98.16)(3,98.84)(4,98.6)(5,95.33)(6,93.59)(7,88.39)(8,80.1)(9, 69.24)};
\addplot[
    color=olive,
    smooth,
    thick,
    mark = triangle*,
    mark size =2pt
    ]
    coordinates {(1,100)(2,100)(3,100)(4,100)(5,100)(6,100)(7,100)(8,100)(9, 100)};
\addplot[
    color=purple,
    smooth,
    thin,
    mark = *,
    mark size =1pt,
    const plot
    ]
    coordinates {(1,100)(2,100)(3,100)(4,100)(5,100)(6,100)(7,100)(8,100)(9, 100)};
\addplot+[
    black,thick,dotted,
    mark=none,
    const plot,
    empty line=jump,
    ]
    coordinates {
    (9,0)
    (9,76.4)
    };

\legend{ClaPIM, EDAM, HD-CAM, Kraken2}
\end{axis}
\end{tikzpicture}
\label{fig:pre_high}
%\vspace{-12pt}
\end{subfigure}%
%%%%%%%%%%%%%%%%%%%%%%%%%%%%%%%%%%%%%%%%%%%%%%%%%%%%%%%%%%%%%%%%%%%%%%%%%%%%%%%%%%%%%%%%%%%%
%%%%%%%%%%%%%%%%%%%%%%%%%%%%%%%%%%%%%%%%%%%%%%%%%%%%%%%%%%%%%%%%%%%%%%%%%%%%%% subfigure 2 
\begin{subfigure}{0.5\textwidth}
\centering 
\textbf{Low Error Rate Reads}\par\medskip
\caption{}%EDAM vs ClaPIM precision for low error rate reads.
\begin{tikzpicture}
\begin{axis}[
    xlabel={Edit Distance Threshold},
    ylabel={Precision(\%)},
    xtick={1,2,3,4,5,6,7,8},
    ytick={75,80,85,90,95,100},
    ymin=75,
    ymax=100,
    legend style={draw=none,fill=none},
    legend pos=south west,
    every axis plot/.append style={ultra thick},
    width=\columnwidth,
    height=0.5\columnwidth
    ]
\addplot[
    color=blue,
    smooth,
    thick,
    mark = *,
    mark size =1pt
    ]
    coordinates {(1,99.5)(2,98.8)(3,97.8)(4,97.1)(5,95.8)(6,94.4)(7,93.27)(8,91.9)};
\addplot[
    color=orange,
    smooth,
    thin,
    mark = *,
    mark size =1pt
    ]
    coordinates {(1,99.74)(2,98.63)(3,96.87)(4,94.42)(5,92.43)(6,88.78)(7,84.99)(8,78.04)};
\addplot[
    color=olive,
    smooth,
    thick,
    mark = triangle*,
    mark size =2pt
    ]
    coordinates {(1,100)(2,99.84)(3,99.58)(4,99.55)(5,99.0)(6,91.3)(7,88.6)	(8,82.21)};
\addplot[
    color=purple,
    smooth,
    thin,
    mark = *,
    mark size =1pt,
    const plot
    ]
    coordinates {(1,98.9)(2,98.9)(3,98.9)(4,98.9)(5,98.9)(6,98.9)(7,98.9)(8,98.9)(9, 98.9)};

\addplot+[
    black,thick,dotted,
    mark=none,
    const plot,
    empty line=jump,
    ]
    coordinates {
    (4,0)
    (4,97.1)
    };

\legend{ClaPIM,EDAM, HD-CAM, Kraken2}
\end{axis}
\end{tikzpicture}
\label{fig:pre_low}
%\vspace{-12pt}
\end{subfigure}

%%%%%%%%%%%%%%%%%%%%%%%%%%%%%%%%%%%%%%%%%%%%%%%%%%%%%%%%%%%%%%%%%%%%%%%%%%%%%%%%%%%%%%%%%%%%
%%%%%%%%%%%%%%%%%%%%%%%%%%%%%%%%%%%%%%%%%%%%%%%%%%%%%%%%%%%%%%%%%%%%%%%%%%%%%% subfigure 3 

\begin{subfigure}{0.5\textwidth}
\centering 
\vspace{10pt}
\caption{}%EDAM vs ClaPIM sensitivity for high error rate reads.
\vspace{-7pt}
\begin{tikzpicture}
\begin{axis}[
    xlabel={Edit Distance Threshold},
    ylabel={Sensitivity(\%)},
    xtick={1,2,3,4,5,6,7,8,9},
    ytick={0,20,40,60,80,100},
    ymin=0,
    ymax=100,
    legend style={draw=none,fill=none},
    legend pos=north west,
    every axis plot/.append style={ultra thick},
    width=\columnwidth,
    height=0.5\columnwidth
    ]
\addplot[
    color=blue,
    smooth,
    thick,
    mark = *,
    mark size =1pt
    ]
    coordinates {(1,4.2)(2,10)(3,17.8)(4,27.4)(5,35)(6,42.1)(7,48.4)(8,56.6)(9, 68.48)};
\addplot[
    color=orange,
    smooth,
    thin,
    mark = *,
    mark size =1pt
    ]
    coordinates {(1,4.72)(2,10.79)(3,16.85)(4,29.03)(5,35.68)(6,43.17)(7,51.15)(8,57.099)(9, 70.23)};
\addplot[
    color=olive,
    smooth,
    thick,
    mark = triangle*,
    mark size =2pt
    ]
    coordinates {(1,0.01)	(2,0.4)	(3,0.59)	(4,0.69)	(5,0.89)	(6,1.1)	(7,1.7)	(8,1.75)	(9,2.28)};
\addplot[
    color=purple,
    smooth,
    thick,
    mark = *,
    mark size =1pt
    ]
    coordinates {(1,1.83)	(2,1.83)	(3,1.83)	(4,1.83)	(5,1.83)	(6,1.83)	(7,1.83)	(8,1.83)	(9,1.83)};
\addplot+[
    black,thick,dotted,
    mark=none,
    const plot,
    empty line=jump,
    ]
    coordinates {
    (9,70.23)
    (9,0)
    };

\legend{ClaPIM, EDAM, HD-CAM, Kraken2}
\end{axis}
\end{tikzpicture}
\label{fig:sen_high}
%\vspace{-12pt}
\end{subfigure}%
%%%%%%%%%%%%%%%%%%%%%%%%%%%%%%%%%%%%%%%%%%%%%%%%%%%%%%%%%%%%%%%%%%%%%%%%%%%%%%%%%%%%%%%%%%%%
%%%%%%%%%%%%%%%%%%%%%%%%%%%%%%%%%%%%%%%%%%%%%%%%%%%%%%%%%%%%%%%%%%%%%%%%%%%%%% subfigure 4 
\begin{subfigure}{0.5\textwidth}
\centering
\vspace{10pt}
\caption{}%EDAM vs ClaPIM sensitivity for low error rate reads.
\vspace{-7pt}
\begin{tikzpicture}
\begin{axis}[
    xlabel={Edit Distance Threshold},
    ylabel={Sensitivity(\%)},
    xtick={1,2,3,4,5,6,7,8},
    ytick={0,20,40,60,80,100},
    ymin=0,
    ymax=100,
    legend style={draw=none},
    legend pos=south east,
    every axis plot/.append style={ultra thick},
    width=\columnwidth,
    height=0.5\columnwidth
    ]
\addplot[
    color=blue,
    smooth,
    thick,
    mark = *,
    mark size =1pt
    ]
    coordinates {(1,67.3)(2,88.9)(3,96.9)(4,98.3)(5,98.5)(6,98.8)(7,98.12)(8,99.2)};
\addplot[
    color=orange,
    smooth,
    thin,
    mark = *,
    mark size =1pt
    ]
    coordinates {(1,73.41)(2,93.39)(3,96.94)(4,99.69)(5,99.71)(6,100)(7,100)(8,100)};
\addplot[
    color=olive,
    smooth,
    thick,
    mark = triangle*,
    mark size =2pt
    ]
    coordinates {(1,12.39)	(2,61.39)	(3,71.63)	(4,77.32)	(5,80.1)	(6,82.07)	(7,82.32)	(8,82.21)};
\addplot[
    color=purple,
    smooth,
    thick,
    mark = *,
    mark size =1pt
    ]
    coordinates {(1,42.8)	(2,42.8)	(3,42.8)	(4,42.8)	(5,42.8)	(6,42.8)	(7,42.8)	(8,42.8)};
\addplot+[
    black,thick,dotted,
    mark=none,
    const plot,
    empty line=jump,
    ]
    coordinates {
    (4,99.69)
    (4,0)
    };

\legend{ClaPIM, EDAM, HD-CAM, Kraken2}
\end{axis}
\end{tikzpicture}
\label{fig:sen_low}
%\vspace{-12pt}
\end{subfigure}

%%%%%%%%%%%%%%%%%%%%%%%%%%%%%%%%%%%%%%%%%%%%%%%%%%%%%%%%%%%%%%%%%%%%%%%%%%%%%%%%%%%%%%%%%%%%
%%%%%%%%%%%%%%%%%%%%%%%%%%%%%%%%%%%%%%%%%%%%%%%%%%%%%%%%%%%%%%%%%%%%%%%%%%%%%% subfigure 5 

\begin{subfigure}{0.5\textwidth}
\centering 
\vspace{10pt}
\caption{}% kraken2 vs ClaPIM F1 score for high error rate reads.
\vspace{-7pt}
\begin{tikzpicture}
\begin{axis}[
    xlabel={Edit Distance Threshold},
    ylabel={F1 score(\%)},
    xtick={1,2,3,4,5,6,7,8,9},
    ytick={0,20,40,60,80,100},
    ymin=0,
    ymax=100,
    legend style={draw=none,fill=none},
    legend pos=north west,
    every axis plot/.append style={ultra thick},
    width=\columnwidth,
    height=0.5\columnwidth
    ]
\addplot[
    color=blue,
    smooth,
    thick,
    mark = *,
    mark size =1pt
    ]
    coordinates {(1,8.06)(2,18.17)(3,30.15)(4,42.87)(5,51.25)(6,58.34)(7,63.67)(8,68.74)(9,72.2)};
\addplot[
    color=orange,
    smooth,
    thick,
    mark = *,
    mark size =1pt
    ]
    coordinates {(1,9)(2,19.4)(3,28.8)(4,44.85)(5,51.9)(6,59.08)(7,64.8)(8,66.67)(9,69.7)};
\addplot[
    color=olive,
    smooth,
    thick,
    mark = triangle*,
    mark size =2pt
    ]
    coordinates {(1,0.02)	(2,0.80)	(3,1.17)	(4,1.37)	(5,1.76)	(6,2.18)	(7,3.34)	(8,3.44)	(9,4.46)};
\addplot[
    color=purple,
    smooth,
    thin,
    mark = *,
    mark size =1pt,
    const plot
    ]
    coordinates {(1,3.6)(2,3.6)(3,3.6)(4,3.6)(5,3.6)(6,3.6)(7,3.6)(8,3.6)(9,3.6)};

\addplot+[
    black,thick,dotted,
    mark=none,
    const plot,
    empty line=jump,
    ]
    coordinates {
    (9,0)
    (9,72.2)
    };

\legend{ClaPIM, EDAM, HD-CAM, Kraken2}
\end{axis}
\end{tikzpicture}
\label{fig:F1_high}
%\vspace{-12pt}
\end{subfigure}%
%%%%%%%%%%%%%%%%%%%%%%%%%%%%%%%%%%%%%%%%%%%%%%%%%%%%%%%%%%%%%%%%%%%%%%%%%%%%%%%%%%%%%%%%%%%%
%%%%%%%%%%%%%%%%%%%%%%%%%%%%%%%%%%%%%%%%%%%%%%%%%%%%%%%%%%%%%%%%%%%%%%%%%%%%%% subfigure 6 
\begin{subfigure}{0.5\textwidth}
\centering 
\vspace{10pt}
\caption{}% kraken2 vs ClaPIM F1 score for low error rate reads.
\vspace{-7pt}
\begin{tikzpicture}
\begin{axis}[
    xlabel={Edit Distance Threshold},
    ylabel={F1 score(\%)},
    xtick={1,2,3,4,5,6,7,8},
    ytick={0,20,40,60,80,100},
    ymin=0,
    ymax=100,
    legend style={draw=none},
    legend pos=south east,
    every axis plot/.append style={ultra thick},
    width=\columnwidth,
    height=0.5\columnwidth
    ]
\addplot[
    color=blue,
    smooth,
    thick,
    mark = *,
    mark size =1pt
    ]
    coordinates {(1,80.29)(2,93.59)(3,97.34)(4,97.7)(5,97.13)(6,96.55)(7,95.63)(8,95.41)};
\addplot[
    color=orange,
    smooth,
    thick,
    mark = *,
    mark size =1pt
    ]
    coordinates {(1,84.57)(2,95.93)(3,96.9)(4,96.9)(5,95.9)(6,94)(7,91.8)(8,87.66)};
\addplot[
    color=olive,
    smooth,
    thick,
    mark = triangle*,
    mark size =2pt
    ]
    coordinates {(1,22.05)	(2,76.03)	(3,83.32)	(4,87.04)	(5,88.75)	(6,86.44)	(7,85.34)	(8,82.21)};
\addplot[
    color=purple,
    smooth,
    thin,
    mark = *,
    mark size =1pt,
    const plot
    ]
    coordinates {(1,60)(2,60)(3,60)(4,60)(5,60)(6,60)(7,60)(8,60)};
\addplot+[
    black,thick,dotted,
    mark=none,
    const plot,
    empty line=jump,
    ]
    coordinates {
    (4,0)
    (4,97.7)
    };

\legend{ClaPIM, EDAM, HD-CAM, Kraken2}
\end{axis}
\end{tikzpicture}
\label{fig:F1_low}
%\vspace{-12pt}
\end{subfigure}

%%%%%%%%%%%%%%%%%%%%%%%%%%%%%%%%%%%%%%%%%%%%%%%%%%%%%%%%%%%%%%%%%%%%%%%%%%%%%%%%%%%%%%%%%%%%
    \caption{Classification quality evaluation. Precision of EDAM, HD-CAM and Kraken2 versus ClaPIM for (a) high and (b) low error rate reads.
    Sensitivity of EDAM, HD-CAM and Kraken2 versus ClaPIM for (c) high and (d) low error rates .
    F1 score of EDAM, HD-CAM and Kraken2 versus ClaPIM for (e) high and (f) low error rate reads.
    The black dotted lines specify the working point of ClaPIM's and EDAM's classification algorithm. The chosen edit distance threshold, which provides the highest F1 score, is $9$ for the high error rate reads and $4$ for the low error rate reads. }
    \label{fig:quality}
\end{figure*}

\subsection{Search Stage Latency}
To assess the searching latency of ClaPIM, we counted and determined the number of cycles needed to perform the classification algorithm described in Section~\ref{sec:Within a Memristive Memory Array}. As parallel in-crossbar and near-crossbar operations are used, the examination of a query can be performed in approximately $6.7\mu s$ (2167 MAGIC clock cycles including initialization cycles, in addition to four SA cycles).

The search stage latency is affected by the number of SA attached to each crossbar. Adding more SAs adds more area overhead to the crossbar and achieves better search latency since more sensing is done in parallel (less iterations for step 5). Table~\ref{tab:tradeoff} presents the trade-off exciting between area overhead and latency for a different number of SAs.

\subsection{Filtering Stage Performance}
To evaluate the filter's run time, we developed an optimized implementation for batching the non-overlapping queries and extracting their lists of crossbars from the tracing table. As mentioned in Section~\ref{sec:Full_Arch}, to ensure the filtering and searching stages are well-balanced, we need to limit the filtering's run time to $6.7 \mu s$ as well. The filtering was executed on an Intel(R) Xeon(R) CPU E5-2683 v4 containing 32 cores, at 2.1 GHz, with 256GB of DRAM, 2400MHz DDR4, 2x 1TB HD, and 480GB SSD, using eight threads. The filter was limited to examining 350 random queries at a time. This limitation resulted in a filtering run time of $6.4 \mu s$, allowing an average of 29 queries to be performed in parallel. Adding the filtering stage to ClaPIM increases its throughput by $29 \times$.

\begin{table}[t]
    \centering
    \caption{Number of SA used: Area vs Latency Trade-off}
    \begin{tabular}{|c|c|c|c|}
        \hline
        \textbf{\#SA} & \textbf{Area overhead} & \textbf{step 5 latency [$ns$]} & \textbf{Total search} \\
         & & & \textbf{stage latency [$\mu s$]} \\
        \hline
        1 & $0\%$ & 4608 & 11.109 \\
        \hline
        2 & $0\%$ & 2304 & 8.805 \\
        \hline
        4 & $1\%$ & 1152 & 7.653 \\
        \hline
        8 & $2\%$ & 576 & 7.077  \\
        \hline
        16 & $4\%$ & 288 & 6.789 \\
        \hline
        32 & $9\%$ & 144 & 6.645 \\
        \hline
        64 & $16\%$ & 72 & 6.573 \\
        \hline
        128 & $28\%$ & 36 & 6.537 \\
        \hline
        
    \end{tabular}
    \label{tab:tradeoff}
    %\vspace{-12pt}
\end{table}

\subsection{The Filter Effect on Energy and Lifetime of the Design}
An in-house simulator\footnote{ https://github.com/marcelkh13/ClaPIM.git } was developed to evaluate the improvement enabled by the filter in energy consumption and the lifetime of the design. The filter was assessed using the following reference DNA: human chromosomes 1, 20, 21, and SARS-CoV-2, downloaded from the National Center for Biotechnology Information (NCBI) online data sets~\cite{NCBI}. On average, less than $0.4\%$ of k-mers had a \textit{potential hit} ($0.4\%$, $0.37\%$, $0.41\%$ and $0.4\%$ for the human chromosomes 1, 20, 21, and SARS-CoV-2, respectively). This leads to at least a $250\times$ reduction in energy consumption compared to the same system without a filtering stage.
Nevertheless, querying in a crossbar involves switching memristor devices, therefore affecting the memory lifetime, given the limited endurance of ReRAM devices~\cite{endurance}. As the filtering drastically reduces the number of crossbars each query should be compared against. Assuming a uniform distribution over the accessed crossbars, a $250\times$ increase in the lifetime of the design is also achieved.  Thus, for a memory lifetime of $10^9$ ($10^{12}$) writes, limited by the endurance of the most frequently accessed cells while assuming wear leveling~\cite{n3xt} (a technique applied to achieve
a uniform distribution of writes in a row). When activating an array to perform a searching task, 7 writes per cell are required. Thus, for the lifetime of the arrays, ClaPIM can perform up to $10^9*250/7= 3.5*10^{10}$ ($10^{12}*250/7 = 3.5*10^{13}$) searching tasks.

%ClaPIM can 
%perform $3.5\cdot10^{10}$ ($3.5\cdot10^{13}$) classification tasks.
%perform $1.9\cdot10^{10}$ ($1.9\cdot10^{13}$) classification tasks.

\subsection{Classification Quality}

\begin{table*}[t]
    \centering
    \begin{threeparttable}
    \caption{Comparison of ClaPIM w/ and w/o filter against other sequence classification tools}
    \label{tab:res}
    \begin{tabular}{|c|c|c|c|c|c|}
        \hline
           & \textbf{Units} & \textbf{ClaPIM (w/o filter)} & \textbf{ClaPIM (w/ filter)} & \textbf{EDAM~\cite{EDAM} \tnote{*}} & \textbf{Kraken2~\cite{Kraken2}} \\
        \hline
        \hline
        \textbf{Throughput} & $G bases/min$ & $0.58$ & $16.82$ & $2,561$  & $9.2$ \\
        \hline
        \textbf{Dynamic Power per Search Against 1 $k$-mer} &  $\mu W$  & $5.7$ & $0.023$ & $60.16$ & - \\
        \hline
        \textbf{Energy per Search Against 1 $k$-mer} & $pJ$ & $37.87$ & $0.15$ & $0.09$ & - \\
        \hline
        \textbf{Density} & $k-mers/\mu m^2$ & $2.17$ & $2.17$ & $0.47\cdot 10^{-3}$ & - \\
        \hline
        \textbf{Area Efficiency = Throughput $\cdot$ Density} & $G bases/min \cdot k-mers/{\mu m^2}$  & $1.26$ & $36.5$ & $1.2$ & - \\
        \hline
        
    \end{tabular}
    \begin{tablenotes}
        \item[*] EDAM's numbers are achieved for small databases ($ \approx 30K $). They do not scale for larger databases.
    \end{tablenotes}
   \end{threeparttable}
\end{table*}

ClaPIM classification quality is compared to (1) EDAM~\cite{EDAM}, an approximates string matching SRAM-based PIM architecture,  (2) Kraken2~\cite{Kraken2}, a state-of-the-art software classifier, and (3) HD-CAM~\cite{HD-CAM}, a Hamming distance tolerant content-addressable memory. The hit confidence numbers from section~\ref{sec:circuit_eval} were integrated with ClaPIM's classification algorithm.

Similar to EDAM and HD-CAM, we evaluated the detection of single species rather than full classification mode. A database containing SARS-CoV-2 (and its variants alpha -- B.1.1.7, beta -- B.1.351, and gamma -- P.1) was used. A synthetic metagenomic sample was created, containing DNA reads of SARS-CoV-2 and its variants and the DNA of several other organisms: SARS-CoV-1, MERSCoV, Coronavirus HKU1 and Human Papillomavirus (HPV) 14.
All the above DNA sequences were downloaded from the NCBI online data sets~\cite{NCBI}.
The 64-base-long DNA reads in the sample were extracted from random positions in the DNA sequences of each of these organisms. Consecutively, sequencing errors (insertions, deletions, and substitutions) were randomly injected, according to two error-rate profiles~\cite{sequencing}: (1) Low error reads of the second generations DNA sequencers (replacement = $3.6\%$, insertion = $0.2\%$, deletion = $0.2\%$) and (2) high error reads of the third generation DNA sequencers (replacement = $1\%$, insertion = $7\%$, deletion = $7\%$).
Fig.~\ref{fig:quality} shows the precision, sensitivity, and F1 of ClaPIM (with the filter) against EDAM, HD-CAM and Kraken2, as a function of the user-defined \textit{edit distance threshold}. This value is used in EDAM's, HD-CAM's, and ClaPIM's classification algorithms, while Kraken2 is unaffected, and thus has a constant value.
The value the user chooses for the edit distance threshold is the one that provides the best classification quality for the algorithm (highest F1 score). The dotted black lines in the graph show the working point of the algorithm (the chosen threshold).
The upper four graphs show that adding the filter improved the precision of the basic algorithm without hurting its sensitivity.
The precision of the classification increased by up to $7\%$ ($3\%$) compared to EDAM for high (low) error rate reads.
Furthermore, compared to Kraken2, as ClaPIM performs approximate $k$-mer matching rather than exact $k$-mer matching, ClaPIM improved the F1 score up to $20\times$ ($1.63\times$) over Kraken2 for the high (low) error profile synthetic reads, as can be seen in the lower two graphs.
Compared to HD-CAM, ClaPIM improves the F1 score by $16.2\times$ ($1.12\times$) for the high (low) error profile reads.
Using the Hamming distance tolerance to perform $k$-mer approximate matching yields good results for low error reads. However, the classification efficiency drops significantly when HD-CAM is applied to high error reads.

\subsection{ClaPIM Throughput and Energy}
Table~\ref{tab:res} summarizes the results for ClaPIM (with and without the filter) and compares them to EDAM and Kraken2.
The results emphasize our design's scalability over EDAM, as we improve the density by $4635\times$. As noted earlier, EDAM does not scale for large databases. Thus, even though it provides impressive results for small databases, it cannot scale to larger, more practical databases. Furthermore, since the periphery used in both ClaPIM and EDAM designs has almost the exact area cost, we infer that under the same area constraints, ClaPIM outperforms EDAM by $30.4\times$, in terms of throughput.
Compared to Kraken2, ClaPIM improves the throughput by $1.8\times$, while, as mentioned before, providing much higher classification quality.
Moreover, as EDAM uses SRAM, additional leakage power is consumed even when searching is not performed, while ClaPIM utilizes nonvolatile devices with approximately zero leakage power.

% ---- Other PIM technologies ---- %
\section{Supporting Other PIM Technologies}
\label{sec:OtherPIM}

% \noindent The same design principles presented in this paper can be implemented using other PIM technologies, such as DRAM-based PIM~\cite{Ambit}. This is possible because all operations, described in the searching stage of ClaPIM (Section \ref{sec:Within a Memristive Memory Array}), are bulk bit-wise operations.
% Mapping the same design to a DRAM-based PIM, for example, might provide lower throughput (higher latency) and lower density but will also eliminate the endurance problem existing in memristors.

The design principles presented in this paper can also be applied to additional PIM technologies beyond memristive PIM. This arises from the fact that ClaPIM is essentially built upon bulk bit-wise logic operations (as described in the searching stage of ClaPIM in Section \ref{sec:Within a Memristive Memory Array}) that are also enabled by other memory technologies.

For example, ClaPIM may also be implemented using DRAM-based PIM such as Ambit~\cite{Ambit}.
Ambit exploits the analog properties of DRAM technology to perform bit-wise operations (AND, OR, and NOT) completely inside DRAM banks. As these gates constitute a functionally-complete set, in-DRAM computation can generalize to any logic function. Therefore, the intra-crossbar searching stage in ClaPIM, which utilizes bulk bit-wise XOR and NOR, may be performed through a sequence of in-DRAM logic operations. For the near-crossbar operations of (1) counting and (2) comparison in ClaPIM, an in-DRAM implementation may perform (1) counting via a sequence of \textit{full adders} and then (2) comparison by subtracting the count result from the pre-defined threshold. The most significant bit of the subtraction result will determine whether the search is a match (as it represents the sign of the difference). While an in-DRAM mapping may result in lower throughput as the counting and comparison stages are performed digitally and not through an efficient analog comparator, it will simultaneously eliminate concerns with memristive PIM, such as endurance.

% ---- Conclusion ---- %
\section{Conclusion}
\label{sec:conclusion}

As sequenced DNA is prone to sequencing errors, performing classification using approximate matching, rather than exact matching, provides significantly higher classification quality. ClaPIM implements a two-stage algorithm -- filtering and searching -- to perform efficient classification based on approximate matching. The filtering stage improves the energy efficiency and the system's lifetime and provides higher classification precision. Additionally, the filtering stage supports a parallel query allocation that increases the overall throughput. In the searching stage, as dense memristive crossbars are used, ClaPIM offers high scalability to support large databases for classification.

Compared with Kraken2, a state-of-the-art CPU baseline classifier, ClaPIM provides significantly higher classification quality and demonstrates a $1.8\times$ improvement in throughput. Compared with HD-CAM, a Hamming distance tolerant content addressable memory, ClaPIM achieves higher classification sensitivity and precision with up to $16.2\times$ higher F1 score. Compared with a recent SRAM-based PIM architecture, EDAM, which is limited to small datasets, ClaPIM provides a $30.4\times$ improvement in area efficiency and provides $7\%$ higher classification precision. 

Furthermore, ClaPIM principles and ideas presented in this paper can be easily modified to be implemented using other PIM technologies, such as DRAM-based PIM~\cite{Ambit}. Designs using other PIM technologies should be studied in the future and compared to determine the best platform.

% ---- References ---- %
\bibliographystyle{IEEEtran}
\bibliography{IEEEabrv,refs}

\vspace*{-0.2cm}

\begin{IEEEbiography}[{\includegraphics[width=1in,height=1.25in,clip,keepaspectratio]{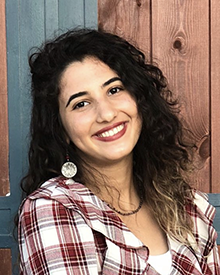}}]{Marcel Khalifa}
is a graduate student at the Andrew and Erna Viterbi Faculty of Electrical and Computer Engineering at the Technion - Israel Institute of Technology. She received the B.Sc. degree in computer engineering, cum laude, in 2020 from the Technion - Israel Institute of Technology. From 2018
to 2020 she was with Intel as a chip designer. Her current
research is focused on accelerating bioinformatics algorithms using processing--in--memory computer architectures with the emerging memory technologies, memristors.
\end{IEEEbiography}
%\vspace{-0.2cm}

\begin{IEEEbiography}[{\includegraphics[width=1in,height=1.25in,clip,keepaspectratio]{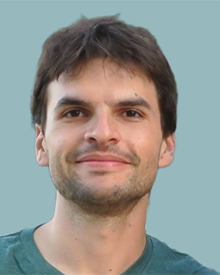}}]{Barak Hoffer}
is a Ph.D. candidate at the Andrew and Erna Viterbi Faculty of Electrical and Computer Engineering, Technion – Israel Institute of Technology.
Barak received the B.Sc. degree in Electrical Engineering in 2018 from the Technion – Israel Institute of Technology.
His main research interests are design and experimental demonstration of processing-in-memory circuits using emerging memory technologies.
\end{IEEEbiography}

\begin{IEEEbiography}[{\includegraphics[width=1in,height=1.25in,clip,keepaspectratio]{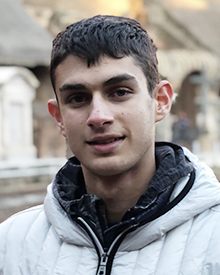}}]{Orian Leitersdorf}
is currently studying towards the B.Sc, M.Sc., and Ph.D. degrees at the Technion – Israel Institute of Technology. He is a scholar at both the Technion Excellence Program and the Technion Lapidim CS Excellence Program, and is also a recipient of the Gutwirth Excellence Scholarship. His current research aims to advance digital processing-in-memory (PIM) towards fundamental applications (e.g., matrix operations, graph algorithms, cryptography) while also addressing challenges such as reliability. Further, his research interests also include several topics in theoretical computer science, such as distributed algorithms and information theory.
\end{IEEEbiography}

\begin{IEEEbiography}[{\includegraphics[width=1in,height=1.25in,clip,keepaspectratio]{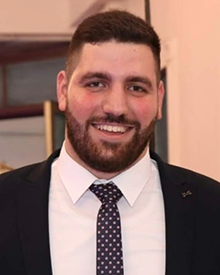}}]{Robert Hanhan}
received the B.Sc. degree in computer engineering from the Technion-Israel Institute of Technology, Haifa, in 2021, where he is currently pursuing the M.Sc. degree in computer engineering. His research interest includes hardware acceleration for genome analysis.
\end{IEEEbiography}

\begin{IEEEbiography}[{\includegraphics[width=1in,height=1.25in,clip,keepaspectratio]{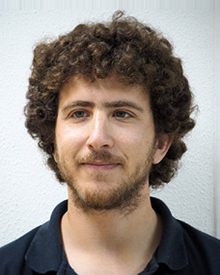}}]{Ben Perach} (Student Member, IEEE)
is a Ph.D. candidate at the Andrew and Erna Viterbi Faculty of Electrical and Computer Engineering, Technion – Israel Institute of Technology. Ben received his B.Sc. degree in mathematics from The Hebrew University of Jerusalem in 2010 and his M.Sc. degree in electrical engineering from Tel Aviv University in 2017. Ben's current research interests include computer architecture with a focus on processor design, and also field-programmable gate arrays, security, and data networks.
\end{IEEEbiography}

\begin{IEEEbiography}[{\includegraphics[width=1in,height=1.25in,clip,keepaspectratio]{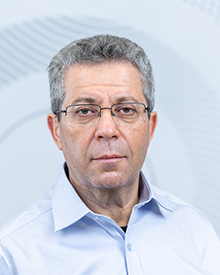}}]{Leonid Yavits}
is with the Faculty of Engineering, Bar-Ilan University, Israel.
He received his MSc and PhD degrees in electrical engineering from Technion, Israel.
Leonid is also a hightech entrepreneur. He co-founded VisionTech, where he co-designed a variety of video compression ICs. VisionTech was acquired by Broadcom in 2001.
Leonid’s research interests include processing in memory, domain specific accelerators, emerging memories, machine learning and bioinformatics.
\end{IEEEbiography}

\begin{IEEEbiography}[{\includegraphics[width=1in,height=1.25in,clip,keepaspectratio]{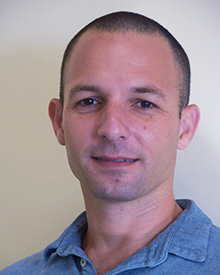}}]{Shahar Kvatinsky}
is an Associate Professor at the Andrew and Erna Viterbi Faculty of Electrical and Computer Engineering, Technion – Israel Institute of Technology. Shahar received the B.Sc. degree in Computer Engineering and Applied Physics and an MBA degree in 2009 and 2010, respectively, both from the Hebrew University of Jerusalem, and the Ph.D. degree in Electrical Engineering from the Technion – Israel Institute of Technology in 2014. From 2006 to 2009, he worked as a circuit designer at Intel. From 2014 to 2015, he was a post-doctoral research fellow at Stanford University. Kvatinsky is a member of the Israel Young Academy. He is the head of the Architecture and Circuits Research Center at the Technion, chair of the IEEE Circuits and Systems in Israel, and an editor of Microelectronics Journal and Array. Kvatinsky has been the recipient of numerous awards: the 2021 Norman Seiden Prize for Academic Excellence, the 2020 MDPI Electronics Young Investigator Award, the 2019 Wolf Foundation's Krill Prize for Excellence in Scientific Research, the 2015 IEEE Guillemin-Cauer Best Paper Award, the 2015 Best Paper of Computer Architecture Letters, Viterbi Fellowship, Jacobs Fellowship, an ERC starting grant, the 2017 Pazy Memorial Award, 2014, 2017 and 2021 Hershel Rich Technion Innovation Awards, the 2013 Sanford Kaplan Prize for Creative Management in High Tech, 2010 Benin prize, and seven Technion excellence teaching awards. His current research is focused on circuits and architectures with emerging memory technologies and the design of energy-efficient architectures.
\end{IEEEbiography}

\end{document}